\documentclass{article} % For LaTeX2e
\usepackage{iclr2026_conference,times}

\usepackage{amsmath,amsfonts,bm}

\def\eqref#1{equation~\ref{#1}}
\def\1{\bm{1}}

\DeclareMathAlphabet{\mathsfit}{\encodingdefault}{\sfdefault}{m}{sl}
\SetMathAlphabet{\mathsfit}{bold}{\encodingdefault}{\sfdefault}{bx}{n}

\usepackage{url}
\usepackage{booktabs} % for table
\usepackage{xcolor}   % for color
\usepackage[table]{xcolor} % for table color
\usepackage{rotating} % for table rotating
\usepackage{adjustbox} % for table rotating
\usepackage{multirow}
\usepackage{multicol}
\usepackage{makecell}
\usepackage{amssymb}  % for \checkmark and so on
\usepackage{bbding} % for \XSolidBrush

\usepackage{algorithm}
\usepackage{algorithmic}

\usepackage{graphicx}    % 基本图形支持
\usepackage{subcaption}  % 子图表支持
\usepackage{tabularx}

\usepackage{nicematrix}

\definecolor{MediumSeaGreen}{RGB}{60, 179, 113}
\definecolor{Pink}{RGB}{255,168,196}
\definecolor{new_orange}{RGB}{208,159,18}
\definecolor{new_blue}{RGB}{52,204,204}
\definecolor{mygray}{RGB}{242,242,242}

\usepackage[colorlinks,linkcolor=red,anchorcolor=blue,citecolor=MediumSeaGreen]{hyperref}

\title{Stability Under Scrutiny: Benchmarking Representation Paradigms for Online HD Mapping}

\author{Hao Shan$^{1,2}$\thanks{Equal contribution.}
\textbf{, Ruikai Li}$^{1,2}$\footnotemark[1]
\textbf{, Han Jiang}$^{1,2}$\thanks{Corresponding author.}
\textbf{, Yizhe Fan}$^{1,2}$
\textbf{, Ziyang Yan}$^{1,2}$
\textbf{, Bohan Li}$^{3,4}$\textbf{,} \\
\textbf{Xiaoshuai Hao}$^{5}$\thanks{Project leader.}
\textbf{, Hao Zhao}$^{5,6}$ 
\textbf{, Zhiyong Cui}$^{1,2}$ 
\textbf{, Yilong Ren}$^{1,2}$
\textbf{, Haiyang Yu}$^{1,2}$ \\
{}$^{1}$ \text{State Key Lab of Intelligent Transportation System, Beihang University} \\ 
{}$^{2}$ \text{School of Transportation Science and Engineering, Beihang University} \\
{}$^{3}$ \text{Shanghai Jiao Tong University} \\
{}$^{4}$ \text{Ningbo Institute of Digital Twin, Eastern Institute of Technology} \\
{}$^{5}$ \text{Beijing Academy of Artificial Intelligence (BAAI)} \\
{}$^{6}$ \text{Institute for AI Industry Research (AIR), Tsinghua University} \\
}

\iclrfinalcopy % Uncomment for camera-ready version, but NOT for submission.
\begin{document}

\maketitle
\vspace{-5mm}
\begin{figure}[htbp]
    \centering
    \includegraphics[width=1.0\linewidth]{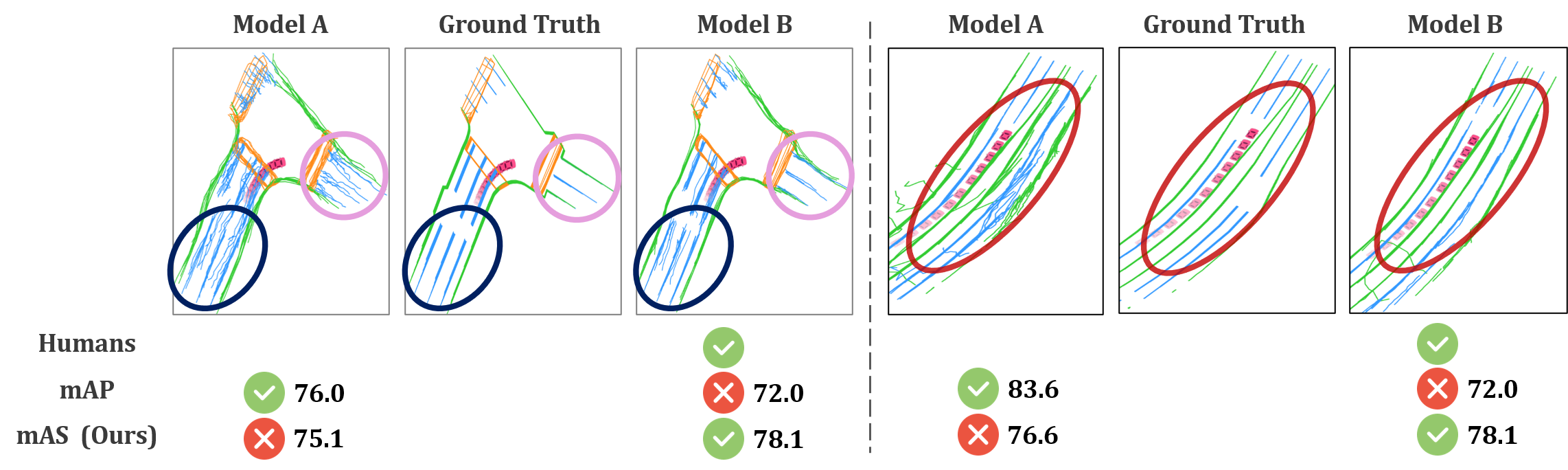}
    \vspace{-5mm}
    \caption{\textbf{Evaluating trustworthiness of online mapping models using human judgment, traditional mAP, and our mAS metric.} In each case, the standard accuracy metric (mAP) fails to align with human judgment because it evaluates only single-frame precision, disregarding stability across time. To address this limitation, we propose the first stability benchmark for online vectorized map construction and present a large-scale analysis of contemporary models.}
    \label{fig:pipeline}
\end{figure}

\begin{abstract}
\vspace{-3mm}
As one of the fundamental modules in autonomous driving, online high-definition (HD) maps have attracted significant attention  due to their cost-effectiveness and real-time capabilities. Since vehicles always cruise in highly dynamic environments, spatial displacement of onboard sensors inevitably causes shifts in real-time HD mapping results, and such instability poses fundamental challenges for downstream tasks. However, existing online map construction models tend to prioritize improving each frame's mapping accuracy, while the mapping stability has not yet been systematically studied. To fill this gap, this paper presents the first comprehensive benchmark for evaluating the temporal stability of online HD mapping models. We propose a multi-dimensional stability evaluation framework with novel metrics for Presence, Localization, and Shape Stability, integrated into a unified mean Average Stability (mAS) score. Extensive experiments on 42 models and variants show that accuracy (mAP) and stability (mAS) represent largely independent performance dimensions. We further analyze the impact of key model design choices on both criteria, identifying architectural and training factors that contribute to high accuracy, high stability, or both. To encourage broader focus on stability, we will release a public benchmark. Our work highlights the importance of treating temporal stability as a core evaluation criterion alongside accuracy, advancing the development of more reliable autonomous driving systems. The benchmark toolkit, code, and models will be available at \hyperlink{https://stablehdmap.github.io/}{https://stablehdmap.github.io/}.
\end{abstract}

\vspace{-3mm}
\section{Introduction}
\label{sec:Introduction}

\begin{figure}[htbp]
    \centering
    \includegraphics[width=1.0\linewidth]{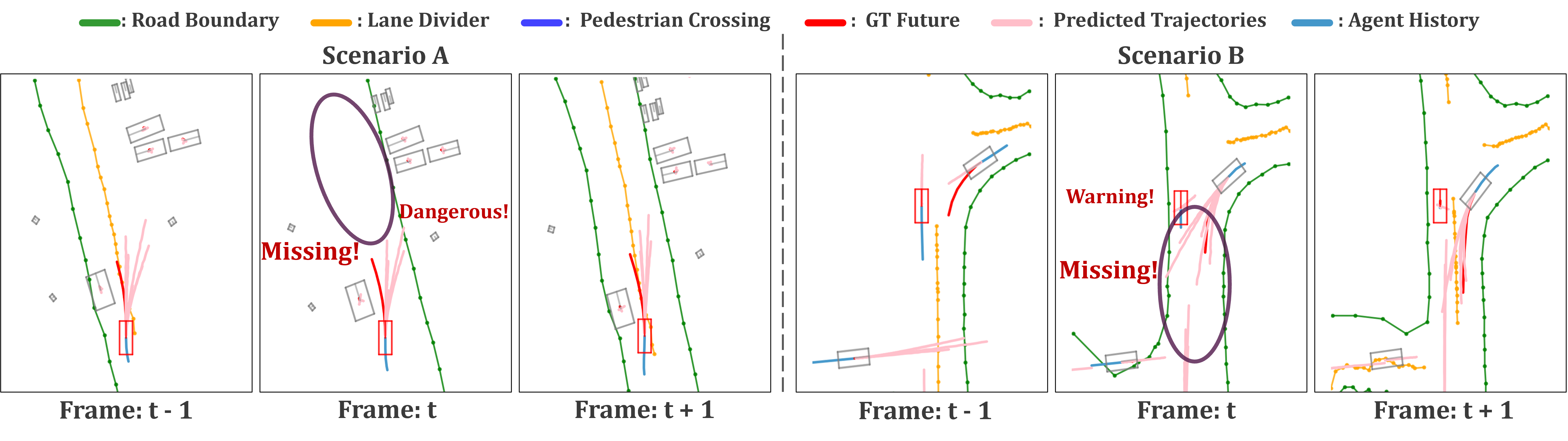}
    \vspace{-5mm}
    \caption{\textbf{The Impact of Unstable Map Elements on Downstream Tasks.} In Scenario A, the ego vehicle attempts to overtake, but the forward lane divider suddenly disappears during the maneuver, causing the ego vehicle to steer toward the curb. In Scenario B, another vehicle attempts to change lanes, but due to flickering lane dividers in the ego vehicle's perception, the ego vehicle interprets the other vehicle's action as a collision course.}
    \label{fig:downstream}
\end{figure}

\vspace{-3mm}
High-definition (HD) map is one of the fundamental component of autonomous driving, offering centimeter-level environmental details such as precise coordinates of map elements and vectorized topological structures \citep{hu2023uniad, jiang2023vad, liao2025diffusiondrive}. Although traditional pre-built HD map provides highly accurate representations, its substantial production and maintenance costs, coupled with limited adaptability to dynamic road conditions, severely restrict large-scale deployment. To address these limitations, online HD mapping has recently emerged as a promising alternative \citep{li2022hdmapnet, liao2022maptr}. By leveraging onboard sensors to perceive the environment in real time, this approach dynamically constructs local vectorized maps, thereby reducing dependence on offline HD maps and paving the way toward scalable and generalizable autonomous driving systems.

Recent advances in online mapping have primarily aimed at improving accuracy and efficiency, giving rise to a diverse set of approaches with distinct representational paradigms \citep{lilja2024localization, liao2022maptr}.% These differences span the entire mapping pipeline: at the input stage, models employ various sensor configurations (e.g., camera-only \citep{zhang2024hrmapnet}, LiDAR-only \citep{wang2023lidar2map}, or camera–LiDAR fusion \citep{liao2025maptrv2}); during feature extraction, 2D backbones (e.g., CNNs \citep{he2016resnet}, ViTs \citep{liu2021swintransformer}) and BEV encoders (e.g., LSS \citep{philion2020lss}, Transformers \citep{chen2022GKT, li2024bevformer}) generate intermediate bird’s-eye-view representations; for temporal modeling, stacking- or streaming-based schemes integrate historical context \citep{liao2022maptr,chen2024maptracker, yuan2024streammapnet}; and at the output stage, decoders with different geometric modeling strategies represent map elements as point sets \citep{liu2024mapqr}, polylines \citep{liu2023vectormapnet}, or parametric curves \citep{qiao2023bemapnet}. 
The community typically evaluates these methods using metrics such as mean Average Precision (mAP) on benchmark datasets, which has driven continuous improvements in state-of-the-art performance. However, a critical yet underexplored issue in traditional evaluation is the stability of model outputs, a property essential for the safe deployment of autonomous driving systems, as illustrated in Fig.\ref{fig:downstream}. A model that achieves high average precision but produces flickering map boundaries or fails entirely at complex intersections, acting like an “intermittently blind” guide, poses substantial safety risks \citep{gu2024traj, zhang2025irostraj}. Despite its importance, the field currently lacks dedicated benchmarks and metrics to quantitatively assess stability in online HD mapping. This gap hinders systematic evaluation of how different representational paradigms respond to real-world disturbances, ultimately slowing progress toward more reliable next-generation mapping systems.

To bridge this gap, we present the first systematic investigation and benchmark for stability in online HD mapping, under the theme “Beyond Accuracy: Under Scrutiny of Stability”. Our key contributions are threefold:

\begin{itemize}
\vspace{-3mm}
\item 
\textbf{A multi-dimensional stability evaluation framework.} We propose novel temporal stability metrics, including Presence, Localization, and Shape Stability, to quantitatively capture the consistency of map elements across consecutive frames. These are integrated into a comprehensive mean Average Stability (mAS) score, enabling holistic model assessment.

\item \textbf{Comprehensive benchmarking and analysis.} We conduct large-scale experiments across diverse state-of-the-art models, revealing that accuracy (mAP) and stability (mAS) are largely independent performance dimensions. Our analysis examines how design choices in sensors, 2D backbones, BEV encoders, temporal fusion, and training regimens influence accuracy and stability as distinct evaluation aspects.

\item \textbf{The first stability centric benchmark.} We establish and will release a public benchmark to catalyze community-wide focus on stability, providing the foundation for developing safer and more robust online mapping systems.
 
\end{itemize}
\vspace{-3mm}
\section{Related Work}
\label{sec:RelatedWork}
\vspace{-3mm}
\paragraph{Online HD Mapping Models.} Online HD mapping has become a critical and extensively studied subtask in autonomous driving. Depending on the choice of sensor input, existing methods can be broadly categorized into camera-only \citep{qiao2023bemapnet, ding2023pivotnet, zhang2023mapvr,liu2024mgmap,liu2024mapqr}, LiDAR-only \citep{wang2023lidar2map}, and camera–LiDAR fusion \citep{li2022hdmapnet,liu2023vectormapnet,liao2022maptr,liao2025maptrv2,yuan2024streammapnet,zhang2024gemap} paradigms, each offering distinct strengths and weaknesses in perception capability and environmental adaptability \citep{hao2024mapdistill,kim2025bridgeta,yan2025mapkd,li2025onestage,kong2025WM}. %At the feature extraction stage, 2D \citep{he2016resnet,liu2021swintransformer} and 3D \citep{lang2019pointpillars,li2024second} backbone networks process images or point clouds to produce hierarchical representations that capture both semantic and geometric cues. For bird’s-eye-view (BEV) feature generation, approaches range from geometry-based \citep{philion2020lss} projection to transformer-based attention mechanisms \citep{li2024bevformer, chen2022GKT}, often incorporating temporal information to enhance coherence across frames \citep{yuan2024streammapnet,chen2024maptracker}. Finally, specialized decoders translate intermediate features into diverse map element representations—such as polylines \citep{li2022hdmapnet, liu2023vectormapnet}, point sets \citep{liao2022maptr,liao2025maptrv2}, or parametric curves \citep{qiao2023bemapnet, ding2023pivotnet}. 
While these paradigms have driven notable progress in mapping accuracy and efficiency, current evaluation frameworks remain narrowly focused on mean Average Precision (mAP), overlooking the critical dimension of stability. This omission substantially limits the practical reliability and deployment of online mapping systems in downstream driving tasks.

\paragraph{Robustness in Autonomous Driving.} Robustness to real-world perturbations has been extensively explored in core autonomous driving tasks. Established benchmarks exist for 2D \citep{wang2020robo2D} or 3D detection \citep{dong2023robo3d1,zhu2023robo3d2,paek2022K-Radar}, segmentation \citep{hong2022roboseg}, and depth estimation \citep{kong2023robodepth}, where models are evaluated under conditions such as corruption, adverse weather, and occlusion. More recently, RoboBEV has extended to Bird’s-Eye-View (BEV) perception, revealing vulnerabilities in view transformation techniques such as LSS and transformers \citep{xie2023robobev,xie2025robobevv2}. In the context of online HD mapping, early efforts have examined sensor level robustness, demonstrating that mapping systems are highly sensitive to corrupted inputs \citep{hao2024your,hao2025msc,hao2025safemap}. However, these studies are restricted to static, single frame analyses and sensor-specific faults. Crucially, the temporal stability of mapping models under sequential perturbations and the comparative robustness of different representation paradigms remain unexplored, a gap which our benchmark aims to address.

\paragraph{Evaluation Metrics for Online HD Mapping.} Current evaluation metrics in the field of online HD map construction are often designed based on single frame geometric accuracy \citep{li2022hdmapnet,liao2022maptr}, primarily focusing on the geometric similarity between the predicted map and ground truth in a given frame. Among typical existing metrics, mean Intersection over Union ($\text{mIoU}$) measures the spatial overlap between the predicted map and the ground truth, while mean Average Precision ($\text{mAP}$) comprehensively considers both classification accuracy and the localization precision of map elements. However, a critical yet previously overlooked issue is that the impact of online mapping on downstream planning tasks depends not only on per-frame geometric accuracy, but also on the inter-frame dynamic stability of the vectorized map. Jitter in map elements across frames can significantly impair the decision-making of autonomous driving systems \citep{zhang2025irostraj,gu2024traj,jiang2023vad}. More seriously, existing metrics completely ignore the temporal geometric stability of map elements, such as the magnitude of polyline edge jitter and the frequency of shape mutations, which are crucial safety factors. To the best of our knowledge, our work is the first to establish a publicly available benchmark dedicated to stability evaluation for online mapping.

\section{Multi-dimensional Map Stability Evaluation Framework}
\label{sec:eval_framework}
\vspace{-3mm}
This section details the proposed framework for multi-dimensional stability evaluation in online HD mapping. The framework quantifies temporal stability through instance-level matching across consecutive frames, specifically designed to assess three critical dimensions: detection consistency, geometric jitter, and shape preservation. The entire pipeline, 
%illustrated in Figure~\ref{fig:pipeline}, 
consists of four main stages:
(1) temporal sampling of frame pairs, (2) cross-frame instance matching, (3) geometric alignment and resampling, and (4) stability metric computation.

\subsection{Temporal Sampling}
\label{subsec:temp_sampling}

\begin{figure}[htbp]
    \centering
    \includegraphics[width=1.00\linewidth]{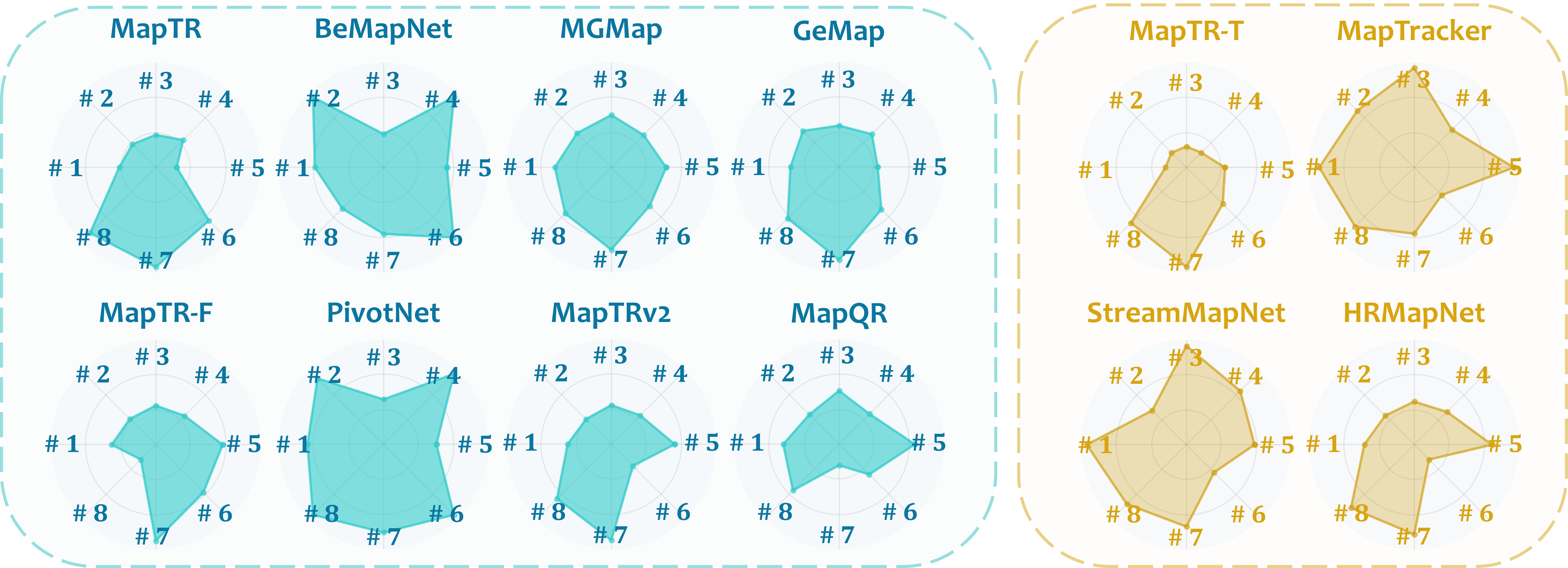}
    \vspace{-5mm}
    \caption{\textbf{Radar chart for Basic HD map constructors covering eight evaluation metrics.} The axes of the radar chart correspond to: \#1 mAS, \#2 Shape, \#3 Loc, \#4 Presence, \#5 mAP, \#6 Inference Memory Cost, \#7 Parameter Count, \#8 FPS.}
    \label{fig:radar_chart}
\end{figure}

The temporal sampling stage constructs pairs of frames for analyzing stability over varying time intervals. Given a sequence of \(L\) consecutive model output frames \(\{D_1, D_2, \ldots, D_L\}\) and a predefined maximum temporal interval \(M\), the process is as follows: for each anchor frame \(D_t\) (where \(t \leq L - M\)), a subsequent frame \(D_{t+k}\) is randomly sampled from the future window \(\{D_{t+1}, \ldots, D_{t+M}\}\), forming an evaluation pair \((D_t, D_{t+k})\). Repeating this procedure for every valid anchor frame \(t\) results in a comprehensive sample set \(S\) of size \(|S| = L - M\), which provides the foundational inputs for subsequent stability analysis.

\subsection{Cross-Frame Instance Matching}
\label{subsec:matching}

Establishing accurate correspondence between map elements across temporal frames is essential for stability assessment. Given the inherent inconsistencies in model predictions, a direct matching approach is prone to error. Instead, a robust indirect strategy is utilized, leveraging the consistent annotations of ground truth (GT) data as a reliable intermediary for association.

For each frame pair \((D_t, D_{t+k})\), the matching process comprises two steps:
\begin{enumerate}
    \item \textbf{Frame-to-GT Matching:} Predictions in each frame are independently matched to their respective GT instances using the Hungarian algorithm, which optimizes a cost function based on geometric and semantic similarity.
    \item \textbf{GT-based Association:} The persistent identification of GT elements across frames enables the linkage of corresponding predictions. Specifically, predictions matched to the same GT instance in different frames are paired, thereby transferring the temporal consistency of the GT to the model outputs.
\end{enumerate}
This procedure yields a set of matched instance pairs \(\{(\text{poly}_{t+k}(e), \text{poly}_{t}(e)) \mid e \in E\}\) for each frame pair, where \(E\) represents the set of successfully tracked map elements. The complete algorithmic details are provided in Algorithm~\ref{alg:cross_frame_matching} of the Appendix.

\subsection{Geometric Alignment and Resampling}
\label{subsec:alignment}

Geometric alignment ensures a fair and spatially consistent comparison between matched polylines \((\text{poly}_{t+k}(e), \text{poly}_{t}(e))\) by transforming them into a common coordinate system and resampling them uniformly. This process consists of three sequential operations.

\noindent\textbf{Coordinate Transformation.} The historical polyline \(\text{poly}_{t}(e)\) is first transformed from the ego coordinate system of its original frame \(D_t\) into the ego coordinate system of the current frame \(D_{t+k}\). This spatial normalization is computed as:
\[
\text{poly}_{t \to t+k}(e) = T_{\text{world} \to t+k} \cdot T_{t \to \text{world}} \cdot \text{poly}_{t}(e),
\]
where \(T_{t \to \text{world}}\) and \(T_{\text{world} \to t+k}\) denote the transformation matrices from frame \(D_t\) to the world frame and from the world frame to frame \(D_{t+k}\), respectively.

\noindent\textbf{Perception Range Filtering.} The transformed polyline \(\text{poly}_{t \to t+k}(e)\) is then clipped to the perception range of the model in frame \(D_{t+k}\). A point \(p = (x, y)\) is retained for subsequent analysis if and only if it satisfies:
\[
x_{\min} \leq x \leq x_{\max}, \quad \text{and} \quad y_{\min} \leq y \leq y_{\max},
\]
where \([x_{\min}, x_{\max}, y_{\min}, y_{\max}]\) defines the operational perceptual boundaries, ensuring evaluation consistency with the model's design.

\noindent\textbf{Uniform Resampling.} Finally, to enable precise point-wise comparison, both the current polyline \(\text{poly}_{t+k}(e)\) and the transformed historical polyline \(\text{poly}_{t \to t+k}(e)\) are resampled along the \(x\)-axis. For their common \(x\)-range \([x_{\min}^p, x_{\max}^p]\), \(N\) equidistant sample points are generated:
\[
\begin{aligned}
x_i &= x_{\min}^p + (i-1) \cdot \frac{x_{\max}^p - x_{\min}^p}{N-1}, \\
\text{poly}_{t+k}^{\text{sample}}(e) &= \{ (x_i, y_{t+k}(x_i)) \mid i = 1, 2, \ldots, N \}, \\
\text{poly}_{t}^{\text{sample}}(e) &= \{ (x_i, y_{t}(x_i)) \mid i = 1, 2, \ldots, N \}.
\end{aligned}
\]
This yields two spatially aligned and uniformly sampled point sets, which serve as the direct input for stability metric computation.

\subsection{Stability Metric Computation}
\label{subsec:3_Metric}
Based on the aligned and resampled point sets \(\text{poly}_{t+k}^{\text{sample}}(e)\) and \(\text{poly}_{t}^{\text{sample}}(e)\), the temporal stability of each matched map element \(e\) is quantified from three perspectives.

\paragraph{Presence Stability.} This metric evaluates the detection consistency of an element across frames. Let \(\text{score}(e)\) denote the model's confidence score for element \(e\) and \(\tau\) be a detection threshold. The presence stability is defined as:
\[
\text{Presence}(e) =
\begin{cases}
1, & \text{if } \text{score}_{t+k}(e) \geq \tau \text{ and } \text{score}_{t}(e) \geq \tau, \\
   & \text{or } \text{score}_{t+k}(e) < \tau \text{ and } \text{score}_{t}(e) < \tau; \\
0.5, & \text{otherwise (flickering)}.
\end{cases}
\]
A higher average value across instances indicates better detection consistency.

\paragraph{Localization Stability.} This metric quantifies the point-wise positional jitter of an element. For the resampled polylines, we compute the average \(L1\) distance in the \(y\)-coordinate and map it to a stability score:
% \[
% \text{Loc}(e) = \exp\left(-\beta \cdot \frac{1}{N} \sum_{i=1}^{N} \left| y_{t+k}(x_i) - y_{t}(x_i) \right| \right),
% \]

\[
\text{Loc}(e) = \beta \cdot \frac{1}{N} \sum_{i=1}^{N} \left| y_{t+k}(x_i) - y_{t}(x_i) \right|,
\]

where \(\beta\) is a scaling parameter. The exponential function translates the average deviation into a score between 0 (unstable) and 1 (stable).

\paragraph{Shape Stability.} This metric assesses the consistency of an element's geometric shape by comparing the curvature of the resampled polylines. We approximate the curvature \(\kappa\) of a polyline as the average angle between consecutive segments:
\[
\kappa(\text{poly}) = \frac{1}{N-1} \sum_{j=1}^{N-1} \theta_j, \quad \text{where} \quad \theta_j = \cos^{-1}\left(\frac{\vec{v_j} \cdot \vec{v_{j+1}}}{|\vec{v_j}| \cdot |\vec{v_{j+1}}|}\right).
\]
The shape stability is then defined as the normalized difference in curvature:
\[
\text{Shape}(e) = 1 - \frac{\left| \kappa(\text{poly}_{t+k}^{\text{sample}}(e)) - \kappa(\text{poly}_{t}^{\text{sample}}(e)) \right|}{\pi}.
\]

\paragraph{Comprehensive Stability Index.} The overall stability for a single instance \(e\) is computed by combining the three metrics:
\[
\text{Stability}(e) = \text{Presence}(e) \cdot \left[ \omega \cdot \text{Loc}(e) + (1 - \omega) \cdot \text{Shape}(e) \right],
\]
where \(\omega \in [0, 1]\) is a weighting parameter (default: 0.7). The class-wise stability is the average over all instances of that class:
\[
\text{Stability}_{\text{class}} = \frac{1}{|\mathcal{I}_{\text{class}}|} \sum_{e \in \mathcal{I}_{\text{class}}} \text{Stability}(e).
\]
Finally, the overall model stability, \textbf{mean Average Stability (mAS)}, is the mean of the stability scores across all classes:
\[
\text{mAS} = \frac{1}{|\mathcal{C}|} \sum_{\text{class} \in \mathcal{C}} \text{Stability}_{\text{class}}.
\]
This single score provides a holistic measure of a model's temporal stability.

\section{Experimental Analysis}
\label{sec:exp}
\vspace{-3mm}
In this section, we present a comprehensive empirical evaluation of our proposed stability assessment framework. Our experiments are designed to answer the following key questions:
\begin{itemize}

\item 
\textbf{RQ1:} How do state-of-the-art online HD mapping models perform in terms of both conventional accuracy (mAP) and our newly proposed temporal stability (mAS)? Is there an implicit correlation between them?

\item 
\textbf{RQ2:} How do different representational paradigms influence model stability?

\item 
\textbf{RQ3:} What are the specific strengths and weaknesses of each paradigm under temporal scrutiny, as revealed by our fine-grained stability metrics (Presence, Localization, Shape)?

\end{itemize}

\begin{table}[htbp]
\centering
%\fontsize{8pt}{10pt}\selectfont

\caption{\textbf{Basic Benchmarking of HD Map Constructors.} Performance comparison of online HD mapping methods on nuScenes val set. Models grouped by temporal fusion mechanisms, input modality, BEV encoder and training epochs. “Temp" denotes the injection of temporal information. “L” and “C” represent LiDAR and camera respectively, while the 2D and 3D backbones employ ResNet50 \citep{he2016resnet} and SECOND \citep{yan2018second}, correspondingly.}
\fontsize{5.5pt}{8pt}\selectfont
\label{tab:main_table}
\begin{tabular}[t]{r|r|cccc|c|ccc|c}
\toprule
\rowcolor{mygray}
\bf Method & 
\bf Venue &
\bf Temp &
\bf Modal &
\bf BEV Encoder &
\bf Epoch &
\bf mAP$\uparrow$ &
\bf Presence$\uparrow$ &
\bf Loc$\uparrow$ &
\bf Shape$\uparrow$ &
\bf \bf mAS$\uparrow$ 
\\
\midrule
\rowcolor{new_blue!3} MapTR \citep{liao2022maptr} & ICLR'23 & \XSolidBrush & C & GKT & 24 & 44.1 & 91.2 & 65.4 &  90.6 & 71.6\\
\rowcolor{new_blue!3} MapTR \citep{liao2022maptr} & ICLR'23 & \XSolidBrush & C \& L & GKT & 24 &  62.8 & 91.5 & 68.6 & 91.0  & 74.0 \\
\rowcolor{new_blue!3} BeMapNet \citep{qiao2023bemapnet} & CVPR'23 & \XSolidBrush & C & IPM-PE & 30 & 61.4 & 100.0 & 65.8 & 97.9 & 81.9\\
\rowcolor{new_blue!3} PivotNet \citep{ding2023pivotnet} & ICCV'23 & \XSolidBrush & C & PersFormer & 30 & 57.1 & 100.0 & 71.4 & 97.2 & 84.3\\
\rowcolor{new_blue!3} MapTRv2 \citep{liao2025maptrv2} & IJCV'24 & \XSolidBrush & C & BEVPool & 24 & 61.4 & 91.5 & 68.6 &  90.9 & 73.9\\
\rowcolor{new_blue!3} GeMap \citep{zhang2024gemap} & ECCV'24 & \XSolidBrush & C & BEVFormer-1 & 24 & 51.3 & 92.3 & 69.7 & 92.6 & 75.5 \\
\rowcolor{new_blue!3} MGMap \citep{liu2024mgmap} & CVPR'24 & \XSolidBrush & C & BEVFormer-1 & 24 & 57.9 & 92.2 & 75.0 & 92.3 & 78.0\\
\rowcolor{new_blue!3} MapQR \citep{liu2024mapqr} & ECCV'24 & \XSolidBrush & C & BEVFormer-3 & 24 & 66.4 & 91.8 & 75.6 & 91.6 & 77.8\\
\midrule
\rowcolor{new_orange!5} MapTR \citep{liao2022maptr} & ICLR'23 & \Checkmark & C & GKT & 24 & 51.3 & 88.61 & 59.7 &  89.3 & 66.6 \\
\rowcolor{new_orange!5} StreamMapNet \citep{yuan2024streammapnet} & WACV'24 & \Checkmark & C & BEVFormer-1 & 30 & 63.3 & 96.6 & 97.7 & 92.3 & 91.9\\
\rowcolor{new_orange!5} MapTracker \citep{chen2024maptracker} & ECCV'24 & \Checkmark & C & BEVFormer-2 & 72 & 75.95 & 93.3 & 98.1  & 95.8 & 90.4\\
\rowcolor{new_orange!5} HRMapNet \citep{zhang2024hrmapnet} & ECCV'24 & \Checkmark & C & BEVFormer-1 & 24 & 67.2 & 92.3 & 70.5 & 91.5 & 75.9\\
\bottomrule
\end{tabular}
\end{table}

\begin{figure}[htbp]
  \centering
  \begin{subfigure}{0.245\textwidth}
    \centering
    \includegraphics[width=\linewidth]{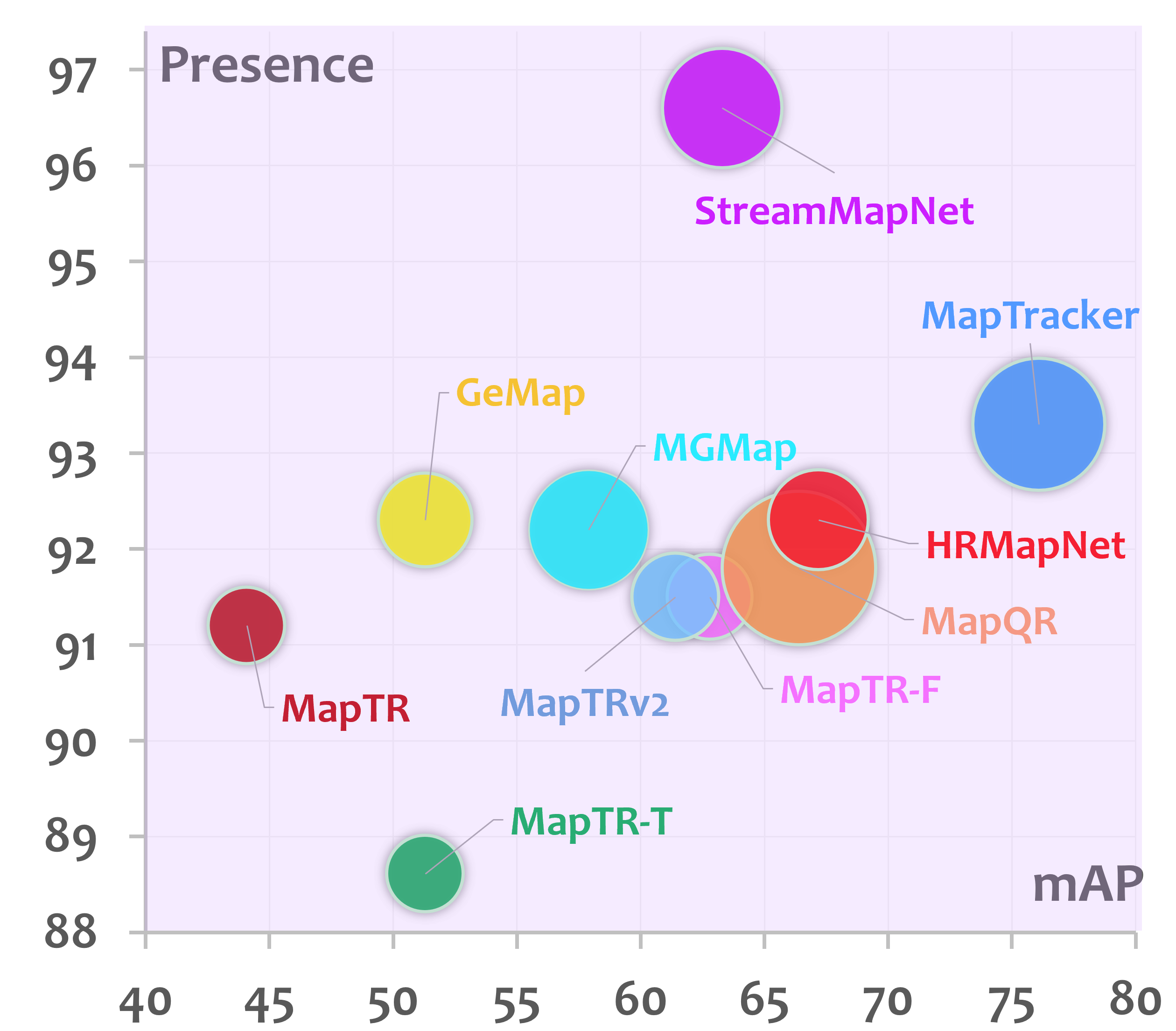}
        \vspace{-5mm}
    \caption{mAP \textit{vs}. Presence}
    \label{fig:rel_sub1}
  \end{subfigure}
  %\hfill
  \begin{subfigure}{0.245\textwidth}
    \centering
    \includegraphics[width=\linewidth]{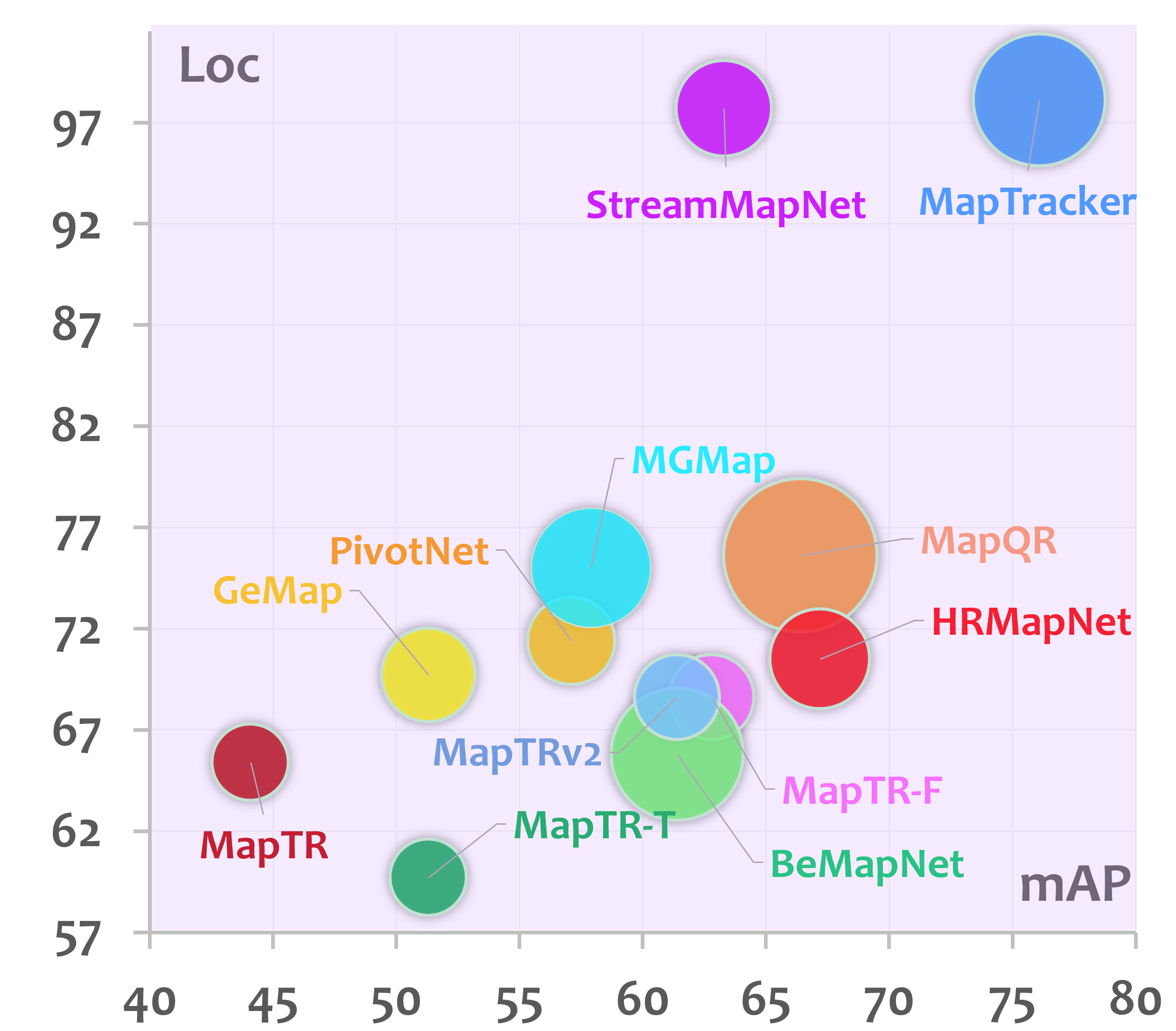}
        \vspace{-5mm}
    \caption{mAP \textit{vs}. Loc}
    \label{fig:rel_sub2}
  \end{subfigure}
  %\hfill
  \begin{subfigure}{0.245\textwidth}
    \centering
    \includegraphics[width=\linewidth]{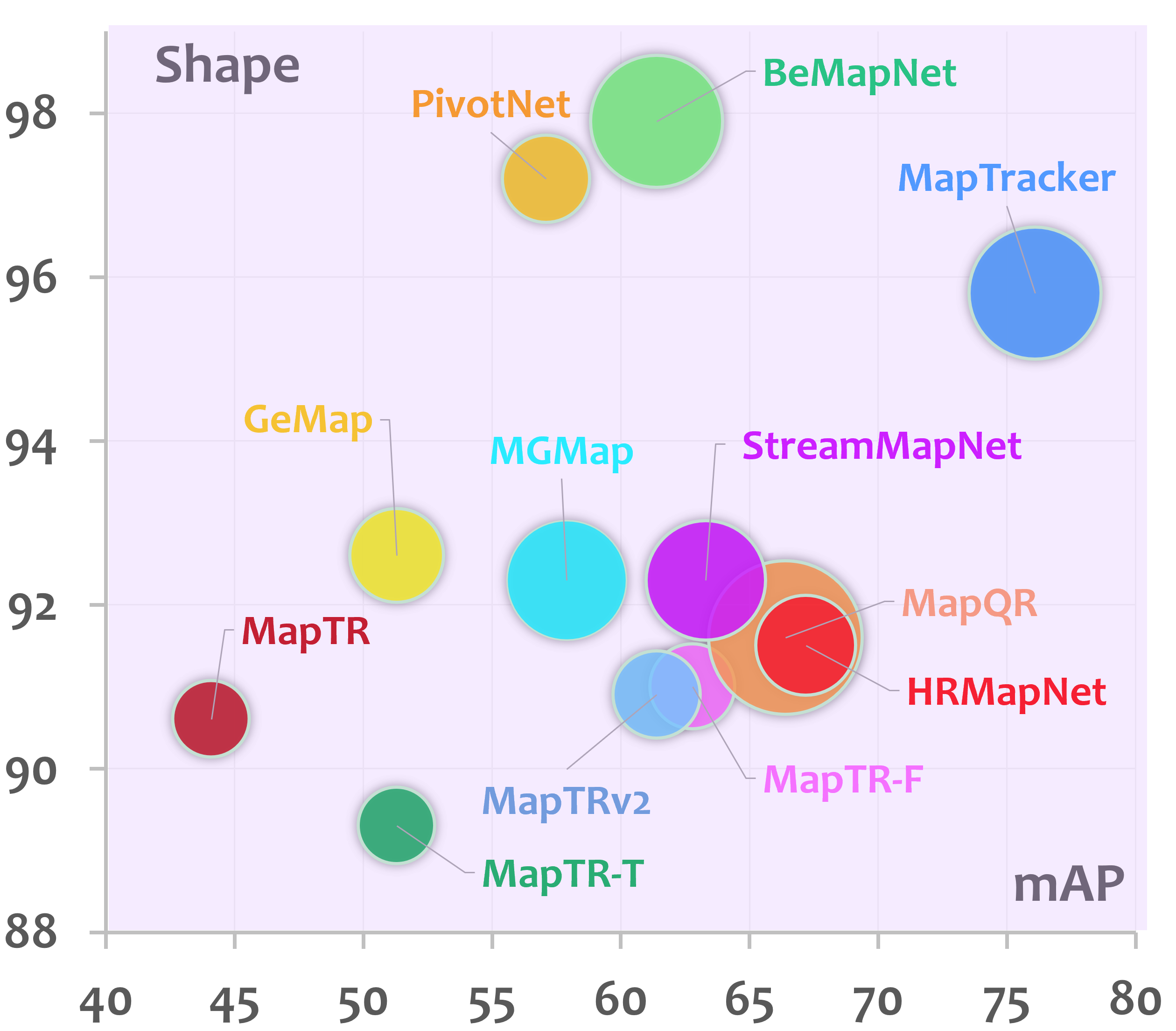}
        \vspace{-5mm}
    \caption{mAP \textit{vs}. Shape}
    \label{fig:rel_sub3}
  \end{subfigure}
  %\hfill
  \begin{subfigure}{0.245\textwidth}
    \centering
    \includegraphics[width=\linewidth]{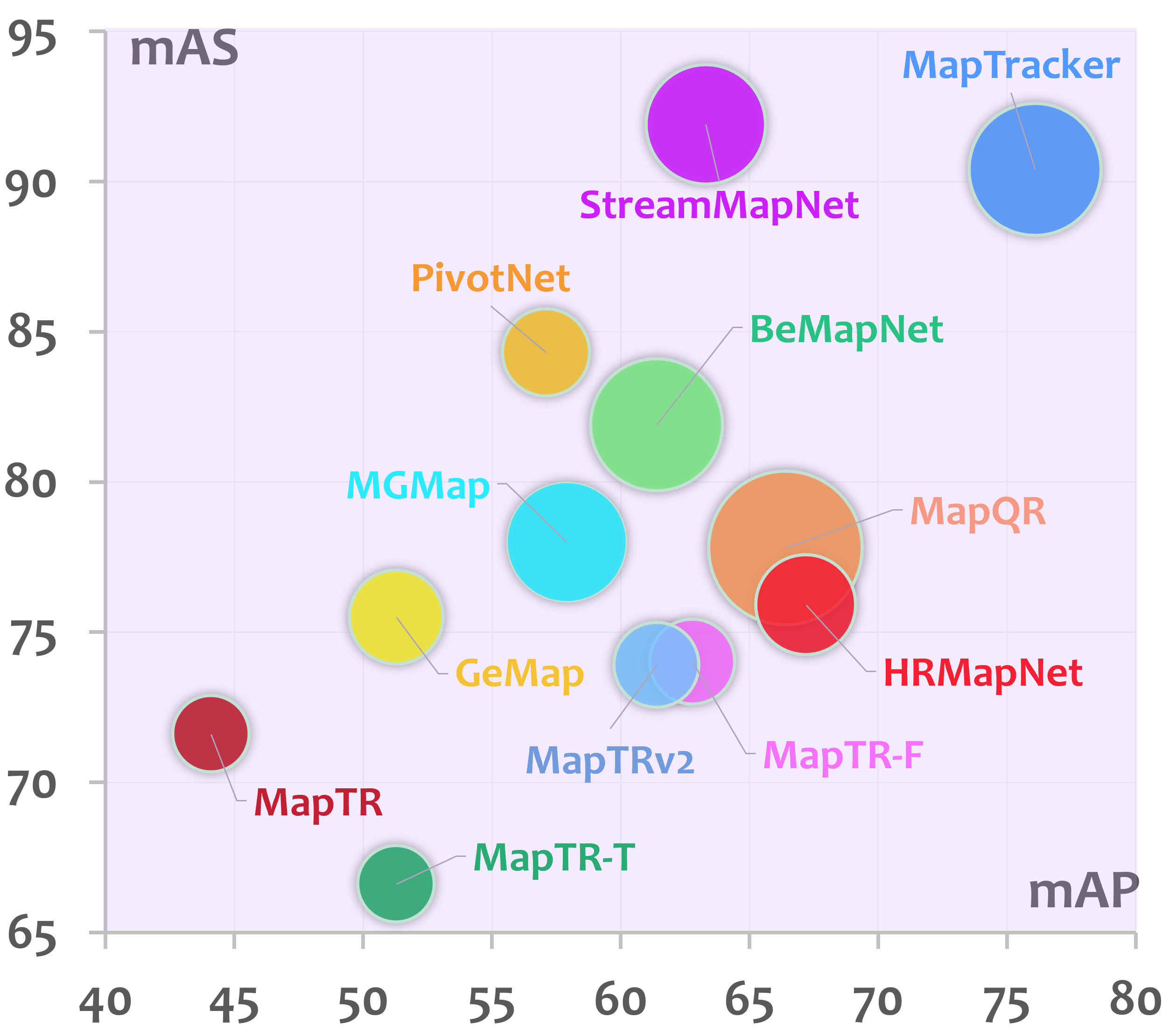}
        \vspace{-5mm}
    \caption{mAP \textit{vs}. mAS}
    \label{fig:rel_sub4}
  \end{subfigure}
  \caption{The correlations between the single-frame accuracy metrics mAP and the stability metrics Presence, Loc, Shape, and mAS. The bubble size represents the model's parameter count.}
  \label{fig:rel_images}
\end{figure}

\subsection{Benchmark Configuration}
\label{subsec:benchmark_config}

\paragraph{Benchmark and Models.} In this work, we conduct a comprehensive evaluation of \textbf{42} online HD map constructors and their variants, covering representative methods following diverse representation paradigms, including BeMapNet \citep{qiao2023bemapnet}, PivotNet \citep{ding2023pivotnet}, MapTR \citep{liao2022maptr}, MapTRv2 \citep{liao2025maptrv2}, StreamMapNet \citep{yuan2024streammapnet}, MGMap \citep{liu2024mgmap}, GeMap \citep{zhang2024gemap}, MapQR \citep{liu2024mapqr}, MapTracker \citep{chen2024maptracker}, and HRMapNet \citep{zhang2024hrmapnet}. These models represent diverse design choices across input modalities, backbone architectures, BEV encoders, temporal fusion mechanisms, historical priors, and training epochs, allowing for a holistic analysis of representational paradigms. Model weights are sourced from official code repositories or retrained using default settings to ensure fairness. Unfortunately, due to the unavailability of source code for several online mapping approaches, we were unable to include them in our full assessment.

\paragraph{Evaluation Metrics.}
We evaluate each model using both the conventional mean Average Precision (mAP) and our proposed multi-dimensional stability metrics: Presence, Localization (Loc), Shape stability, and the comprehensive mean Average Stability (mAS), as defined in Section \ref{subsec:3_Metric}. Additional metrics related to inference performance have been incorporated into the evaluation framework, As shown in Fig.\ref{fig:radar_chart} More detailed evaluation metrics are provided in Section \ref{sec:appendix_6_model_details} of the Appendix.

\subsection{Basic Performance Benchmarking (RQ1)}

The basic benchmarking results are summarized in Table \ref{tab:main_table}. Our analysis reveals two key findings that challenge the sole reliance on accuracy for model evaluation:

\paragraph{Stability constitutes a distinct and critical performance dimension.} A primary observation from our benchmark, as illustrated in Fig.\ref{fig:rel_images}, is the imperfect correlation between conventional accuracy (mAP) and temporal stability (mAS). We observe that models with higher mAP do not necessarily achieve superior mAS, indicating that temporal stability is not an automatic byproduct of high accuracy but rather a unique aspect of model performance. This aspect is crucial for real-world deployment yet is overlooked by conventional metrics.

\paragraph{Significant stability gaps exist among mainstream paradigms.} We observe that the mAS scores span a wide range from 71.6 (MapTR \citep{liao2022maptr}) to 91.9 (StreamMapNet \citep{yuan2024streammapnet}), indicating that the choice of representational paradigm profoundly impacts the consistency of the generated map. A majority of existing models, cluster in the lower to mid-range of mAS (71.6 - 78.0). This clustering suggests a common challenge faced by current approaches in maintaining output stability across consecutive frames.

\subsection{In-depth Analysis of Representational Paradigms (RQ2 \& RQ3)}
\begin{table}[htbp]
\centering
% 第一个表格
\begin{minipage}[t]{0.48\textwidth}
\centering
\caption{Ablation on the Input Modality.}
\fontsize{5.5pt}{8pt}\selectfont
\label{tab:ablation_modal}
\vspace{-\baselineskip} % 调整标题间距
\begin{NiceTabular}{@{\hspace{0.5em}}r!{\vrule width 0.2pt}c!{\vrule width 0.2pt}c!{\vrule width 0.2pt}ccc!{\vrule width 0.2pt}c@{\hspace{0.5em}}}
\toprule
\rowcolor{mygray}
\bf Method & \bf Modal & 
\bf mAP &
\bf Presence &
\bf Loc &
\bf Shape &
\bf mAS   \\
\midrule
MapTR \textcolor{new_orange}{$\circ$}& C & 44.1 & 91.2 & 65.4 &  90.6 & 71.6  \\
\rowcolor{new_orange!15} MapTR \textcolor{new_orange}{$\bullet$}  & C \& L & \textbf{62.8} & \textbf{91.5} & \textbf{68.6} & \textbf{91.0}  & \textbf{74.0}  \\
GeMap \textcolor{new_orange}{$\circ$} & C & 62.7 & \textbf{91.1} & \textbf{67.5} & \textbf{94.5} & \textbf{74.7} \\
\rowcolor{new_orange!12} GeMap \textcolor{new_orange}{$\bullet$} & C \& L & \textbf{66.5} & 89.1 & 66.3 &  92.7 & 71.8 \\
\bottomrule
\end{NiceTabular}
\end{minipage}
\hspace{0.005\textwidth} % 控制两个表格之间的间距
% 第二个表格
\begin{minipage}[t]{0.49\textwidth}
\centering
\caption{Ablation on the BEV Encoder.}
\fontsize{5.5pt}{8pt}\selectfont
\label{tab:ablation_bev}
\vspace{-\baselineskip} % 调整标题间距
\begin{NiceTabular}{@{\hspace{0.5em}}r!{\vrule width 0.2pt}c!{\vrule width 0.2pt}c!{\vrule width 0.2pt}ccc!{\vrule width 0.2pt}c@{\hspace{0.5em}}}
\toprule
\rowcolor{mygray}
\bf Method & \bf Encoder & 
\bf mAP &
\bf Presence &
\bf Loc &
\bf Shape &
\bf mAS   \\
\midrule
MapTR \textcolor{new_blue}{$\circ$} & BEVFormer & 41.6 & 89.6 & 69.7 & \textbf{90.6} & 71.3  \\
MapTR \textcolor{new_blue}{$\circ$} & GKT & 44.1 & \textbf{91.2} & 65.4 & \textbf{90.6} & 71.6 \\
\rowcolor{new_blue!15} MapTR \textcolor{new_blue}{$\bullet$} & BEVPool & \textbf{50.1} & 89.3 & \textbf{69.8} & 88.5 & \textbf{71.9} \\
\bottomrule
\end{NiceTabular}
\end{minipage}
\end{table}

\paragraph{Impact of Sensor Modality.} Our analysis reveals a nuanced relationship between sensor modality and temporal stability. As shown in Table 2, while LiDAR fusion consistently improves perception accuracy, increasing MapTR's \citep{liao2022maptr} mAP by 42.6\% (from 44.1 to 62.8) and GeMap's \citep{zhang2024gemap} mAP by 6.1\% (from 62.7 to 66.5), its effect on temporal stability demonstrates significant model dependence.  MapTR \citep{liao2022maptr} benefits from sensor fusion with a 3.4\% improvement in mAS (71.6 to 74.0), suggesting that LiDAR's precise depth measurements can enhance temporal consistency. In contrast, GeMap \citep{zhang2024gemap} experiences a 3.9\% decrease in mAS (74.7 to 71.8) despite accuracy gains, indicating potential architectural limitations in leveraging multi-modal signals for stable predictions. This divergence highlights that additional sensors alone cannot guarantee improved stability.

\paragraph{Influence of BEV Encoding Strategies.} Our ablation study on MapTR \citep{liao2022maptr} demonstrates that different BEV encoders achieve similar overall temporal stability, with mAS scores ranging from 71.3 to 71.9, despite significant variations in accuracy, where mAP values span from 41.6 to 50.1, as summarized in Table \ref{tab:ablation_bev}. Further analysis reveals distinct specialization patterns among encoders. The GKT \citep{chen2022GKT} encoder achieves superior Presence Stability at 91.2, ensuring consistent detection of map elements across frames. In comparison, BEVFormer \citep{li2024bevformer} and BEVPool \citep{liu2022bevfusion} excel in Localization Stability, with scores of 69.7 and 69.8 respectively, indicating their stronger capability in mitigating geometric jitter. These results highlight that BEV encoders embody characteristic preferences for different aspects of temporal stability, even within the same model architecture. % \TODO{XXXXXX} This indicates that the encoder selection should be guided by the specific stability priority of the application.

\begin{table}[htbp]
\centering
%\fontsize{8pt}{10pt}\selectfont

\caption{Ablation on Temporal Fusion.}
\fontsize{6pt}{8pt}\selectfont
\label{tab:ablation_temporal}
\begin{tabular}[t]{r|cc|ccc|c|ccc|c}
\toprule
\rowcolor{mygray}
\bf Method & 
\bf Temp &
\bf Initial Map &
\bf Back. &
\bf BEV Encoder &
\bf Epoch &
\bf mAP$\uparrow$ &
\bf Presence$\uparrow$ &
\bf Loc$\uparrow$ &
\bf Shape$\uparrow$ &
\bf \bf mAS$\uparrow$ 
\\
\midrule
MapTR \textcolor{new_orange}{$\circ$} & \XSolidBrush & \XSolidBrush & R50 & GKT & 24 & 44.1 & \textbf{91.2} & \textbf{65.4} &  \textbf{90.6} & \textbf{71.6} \\
\rowcolor{new_orange!5} MapTR \textcolor{new_orange}{$\bullet$} & \Checkmark & \XSolidBrush & R50 & GKT & 24 & \textbf{51.3} & 88.6 & 59.7 &  89.3 & 66.6 \\
MapTR \textcolor{new_orange}{$\circ$}  & \XSolidBrush & \XSolidBrush & R50 & BEVFormer & 24 & 41.6 & 89.6 & \textbf{69.7} &  90.6 & 71.3 \\
\rowcolor{new_orange!5} MapTR \textcolor{new_orange}{$\bullet$}  & \Checkmark & \XSolidBrush & R50 & BEVFormer & 24 & \textbf{53.3} & \textbf{90.4} & 69.5 &  \textbf{91.2} & \textbf{73.0} \\
StreamMapNet \textcolor{new_orange}{$\circ$}  & \XSolidBrush & \XSolidBrush & R50 & BEVFormer-1 & 30 & 51.7 & 87.0 & \textbf{97.8} &  \textbf{95.1} & 83.8 \\
\rowcolor{new_orange!5} StreamMapNet \textcolor{new_orange}{$\bullet$}  & \Checkmark & \XSolidBrush & R50 & BEVFormer-1 & 30 & \textbf{63.3} & \textbf{96.6} & 97.7 & 92.3 & \textbf{91.9} \\
MapTracker \textcolor{new_orange}{$\circ$}  & \XSolidBrush & \XSolidBrush & R18 & BEVFormer-2 & 72 & 62.8 & \textbf{95.3} & 97.3 & 85.9  & 87.4 \\
\rowcolor{new_orange!5} MapTracker \textcolor{new_orange}{$\bullet$} & \Checkmark & \XSolidBrush & R18 & BEVFormer-2 & 72 & \textbf{71.9} & 92.9 & \textbf{98.5}  & \textbf{94.8} & \textbf{89.9} \\
MapTracker \textcolor{new_orange}{$\circ$} & \XSolidBrush & \XSolidBrush & R50 & BEVFormer-2 & 72 & 68.3 & \textbf{94.5} & 97.9  & 93.8 & \textbf{90.8} \\
\rowcolor{new_orange!5} MapTracker \textcolor{new_orange}{$\bullet$} & \Checkmark & \XSolidBrush & R50 & BEVFormer-2 & 72 &\textbf{ 75.95} & 93.3 & \textbf{98.1}  & \textbf{95.8} & 90.4 \\
HRMapNet \textcolor{new_orange}{$\circ$} & \Checkmark & \XSolidBrush & R50 & BEVFormer-1 & 24 & 67.2 & 92.3 & 70.5 & 91.5 & 75.9 \\
HRMapNet \textcolor{new_orange}{$\circ$} & \Checkmark & Testing Map & R50 & BEVFormer-1 & 24 & 73.0 & \textbf{94.9} & 71.4 & 93.0 & \textbf{78.4} \\
\rowcolor{new_orange!5} HRMapNet \textcolor{new_orange}{$\bullet$} & \Checkmark & Training Map & R50 & BEVFormer-1 & 24 & \textbf{83.6} & 89.9 & \textbf{75.9} & \textbf{93.2} & 76.7 \\

\bottomrule
\end{tabular}
\end{table}
\begin{figure}[htbp]
    \centering
    \includegraphics[width=0.98\linewidth]{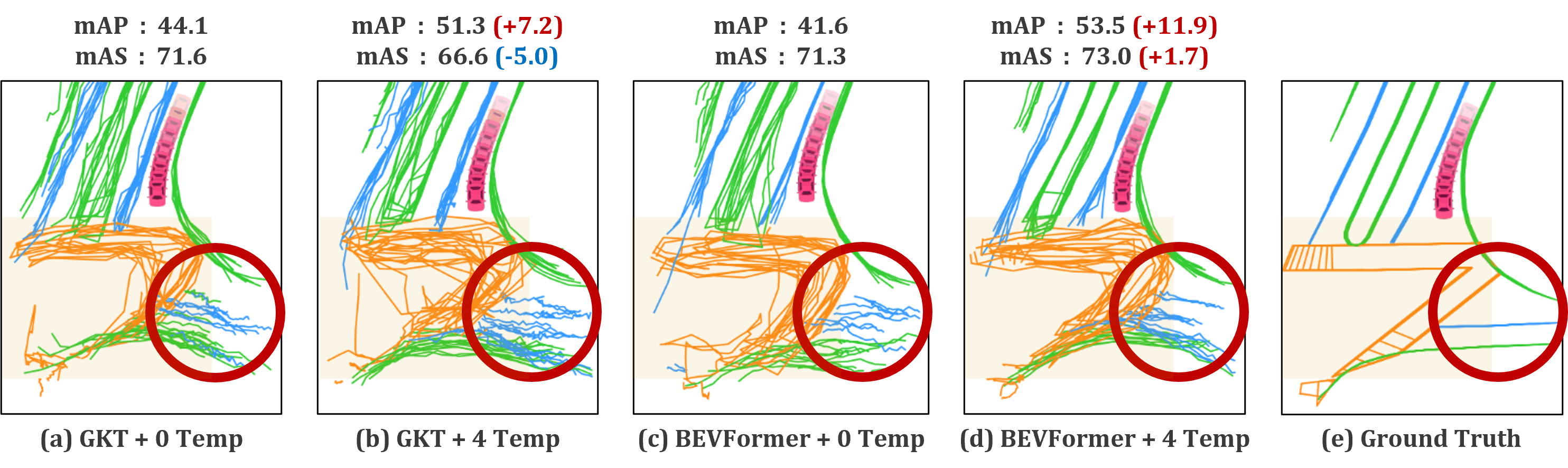}
        \vspace{-3mm}
    \caption{The dual effect of temporal fusion on MapTR with different BEV encoders.}
    \label{fig:vis_maptr_temporal}
\end{figure}

\paragraph{Discussion of Temporal Fusion.} As shown in Table \ref{tab:ablation_temporal}, the effectiveness of temporal fusion is highly dependent on architectural compatibility. Models with native temporal designs demonstrate robust performance: StreamMapNet \citep{yuan2024streammapnet} achieves exceptional temporal stability (mAS: 91.9), while MapTracker \citep{chen2024maptracker} maintains strong stability (mAS: 90.4) alongside significant mAP improvement (+11.4\%). In contrast, adding temporal fusion to architectures not originally designed for temporal processing yields inconsistent results. MapTR \citep{liao2022maptr} exhibits divergent behaviors depending on its BEV encoder. With the GKT \citep{chen2022GKT} encoder, temporal fusion degrades stability (mAS: -7.0\%), whereas with BEVFormer \citep{li2024bevformer}, it provides balanced improvement (mAS: +2.4\%, mAP: +28.1\%). This contrast highlights the critical influence of the encoder's representation capacity on temporal integration. Furthermore, HRMapNet \citep{zhang2024hrmapnet} demonstrates that while map priors substantially boost accuracy (mAP: +24.4\% with training map priors), their impact on stability is more limited (mAS: +1.1\%). This suggests that dynamic temporal modeling contributes more significantly to consistency than static priors. These findings collectively emphasize that effective temporal fusion requires co-design of architectural components rather than simply appending temporal modules.

\begin{table}[htbp]
\centering
% 第一个表格
\begin{minipage}[t]{0.48\textwidth}
\centering
\caption{Ablation on the 2D Backbone.}
\fontsize{5.5pt}{8pt}\selectfont
\label{tab:ablation_2D_Backbone}
\vspace{-\baselineskip} % 调整标题间距
\begin{NiceTabular}{@{\hspace{0.5em}}r!{\vrule width 0.2pt}c!{\vrule width 0.2pt}c!{\vrule width 0.2pt}ccc!{\vrule width 0.2pt}c@{\hspace{0.5em}}}
\toprule
\rowcolor{mygray}
\bf Method & \bf Back. & 
\bf mAP &
\bf Presence &
\bf Loc &
\bf Shape &
\bf mAS   \\
\midrule
MapTR \textcolor{new_orange}{$\circ$}& R18 & 32.4 & 87.8 & \textbf{75.0} & 88.5  &  \textbf{72.8} \\
\rowcolor{new_orange!15} MapTR \textcolor{new_orange}{$\bullet$}& R50  & \textbf{44.1} & \textbf{91.2} & 65.4 & \textbf{90.6}  & 71.6  \\
MapTRv2 \textcolor{new_orange}{$\circ$}& R18 & 57.2  & 91.0 & \textbf{73.2} & \textbf{91.2}  & \textbf{75.6}  \\
\rowcolor{new_orange!15} MapTRv2 \textcolor{new_orange}{$\bullet$}& R50 & \textbf{61.4}  & \textbf{91.5} & 68.6 & 91.0  &  74.0 \\
MapQR \textcolor{new_orange}{$\circ$}& R18 & 62.3  & 88.2 & 73.1 & \textbf{92.5}  & 74.1  \\
\rowcolor{new_orange!15} MapQR \textcolor{new_orange}{$\bullet$}& R50 & \textbf{66.4}  & \textbf{91.8} & \textbf{75.6} & 91.6  & \textbf{77.8} \\
BeMapNet \textcolor{new_orange}{$\circ$}& Effi-B0 & 60.7  & \textbf{100.0} & \textbf{67.9} &  97.9 & \textbf{82.9}  \\
BeMapNet \textcolor{new_orange}{$\circ$}& R50 & 63.6  & \textbf{100.0} & 65.8  & 97.9  &  81.9 \\
\rowcolor{new_orange!15} BeMapNet \textcolor{new_orange}{$\bullet$}& Swin-T & \textbf{64.1} & \textbf{100.0} & 62.8 & \textbf{98.0}  & 80.4  \\
PivotNet \textcolor{new_orange}{$\circ$}& Effi-B0 & 57.8  & \textbf{100.0} & \textbf{71.8} & \textbf{97.2}  & \textbf{84.5}  \\
PivotNet \textcolor{new_orange}{$\circ$}& R50 & 57.1  & \textbf{100.0} & 71.4 & \textbf{97.2}  &  84.3 \\
\rowcolor{new_orange!15} PivotNet \textcolor{new_orange}{$\bullet$}& Swin-T & \textbf{61.6}  & \textbf{100.0} & 71.6 & \textbf{97.2}  & 84.4  \\
GeMap \textcolor{new_orange}{$\circ$}& R50 & 62.7  & 91.1 & 67.5 & \textbf{94.5}  & 74.7  \\
GeMap \textcolor{new_orange}{$\circ$}& Swin-T & 72.0  & 92.2 & \textbf{74.9} & 93.3  & \textbf{78.1}  \\
GeMap \textcolor{new_orange}{$\circ$}& V2-99 & 72.0  & 89.2 & 71.5 &  82.6 &  74.2 \\
\rowcolor{new_orange!15} GeMap \textcolor{new_orange}{$\bullet$}& V2-99* & \textbf{76.0}  & \textbf{93.4} & 67.0 & 93.7  & 75.1  \\
\bottomrule
\end{NiceTabular}
\end{minipage}
\hspace{0.005\textwidth} % 控制两个表格之间的间距
% 第二个表格
\begin{minipage}[t]{0.49\textwidth}
\centering
\caption{Ablation on the Training Epochs.}
\fontsize{5.5pt}{8pt}\selectfont
\label{tab:ablation_epoch}
\vspace{-\baselineskip} % 调整标题间距
\begin{NiceTabular}{@{\hspace{0.5em}}r!{\vrule width 0.2pt}c!{\vrule width 0.2pt}c!{\vrule width 0.2pt}ccc!{\vrule width 0.2pt}c@{\hspace{0.5em}}}
\toprule
\rowcolor{mygray}
\bf Method & \bf Epoch & 
\bf mAP &
\bf Presence &
\bf Loc &
\bf Shape &
\bf mAS   \\
\midrule
MapTR-18 \textcolor{new_blue}{$\circ$} & 24 & 32.4 & \textbf{87.8} & \textbf{75.0} & 88.5 &  \textbf{72.8} \\
\rowcolor{new_blue!15} MapTR-18 \textcolor{new_blue}{$\bullet$} & 110 & \textbf{45.5} & 86.0 & 71.7 & \textbf{94.8} & 71.9  \\
MapTR-50 \textcolor{new_blue}{$\circ$} & 24 & 44.1 & \textbf{91.2} & \textbf{65.4} & \textbf{90.6} &  \textbf{71.6} \\
\rowcolor{new_blue!15} MapTR-50 \textcolor{new_blue}{$\bullet$} & 110 & \textbf{50.5} & 89.8 & 63.2 & 91.0 &  68.2 \\
MapQR-50 \textcolor{new_blue}{$\circ$} & 24 & 66.4 & 91.8 & 75.6 & 91.6 &  77.8 \\
\rowcolor{new_blue!15} MapQR-50 \textcolor{new_blue}{$\bullet$} & 110 & \textbf{72.6} & \textbf{92.4} & \textbf{75.9} & \textbf{96.4} &  \textbf{80.3} \\
GeMap-50 \textcolor{new_blue}{$\circ$} & 24 & 51.3 & \textbf{92.3} & \textbf{69.7} & 92.6 & \textbf{75.5}  \\
\rowcolor{new_blue!15} GeMap-50 \textcolor{new_blue}{$\bullet$} & 110 & \textbf{62.9} & 91.1 & 67.5 & \textbf{94.5} & 74.7  \\
BeMapNet \textcolor{new_blue}{$\circ$} & 30 & 64.1 & \textbf{100.0} & 62.8 & 98.0  & 80.4  \\
\rowcolor{new_blue!15} BeMapNet \textcolor{new_blue}{$\bullet$} & 110 & \textbf{68.3} & \textbf{100.0} & \textbf{64.0} & \textbf{98.2} &  \textbf{81.1} \\
PivotNet \textcolor{new_blue}{$\circ$} & 30 & 61.6 & \textbf{100.0} & 71.6 & 97.2 & 84.4  \\
\rowcolor{new_blue!15} PivotNet \textcolor{new_blue}{$\bullet$} & 110 & \textbf{66.4} & \textbf{100.0} & \textbf{72.1} & \textbf{97.4} & \textbf{84.8}  \\
MapTracker-18 \textcolor{new_blue}{$\circ$} & 48 & 69.3 & \textbf{94.8} & 98.2 & 93.8 & \textbf{90.8}  \\
\rowcolor{new_blue!15} MapTracker-18 \textcolor{new_blue}{$\bullet$} & 72 & \textbf{71.9} & 92.9 & \textbf{98.5} & \textbf{94.8} & 89.9  \\
MapTracker-50 \textcolor{new_blue}{$\circ$} & 48 & 72.96 & 91.8  & \textbf{98.5} & \textbf{96.0} &  \textbf{91.7} \\
\rowcolor{new_blue!15} MapTracker-50 \textcolor{new_blue}{$\bullet$} & 72 & \textbf{75.95} & \textbf{93.3} & 98.1 & 95.8 & 90.4  \\
\bottomrule
\end{NiceTabular}
\end{minipage}
\end{table}

\paragraph{Influence of The 2D Backbone.} The impact of the 2D backbone is model-specific as indicated in Table \ref{tab:ablation_2D_Backbone}. A more powerful backbone consistently improves accuracy (mAP), as seen in MapTR (+36.1\%) \citep{liao2022maptr} and MapQR (+6.6\%) \citep{liu2024mapqr}. However, its effect on stability (mAS) is less predictable, ranging from a slight decrease in MapTR (-1.6\%) \citep{liao2022maptr} to an increase in MapQR (+5.0\%) \citep{liu2024mapqr}. We observe a recurring trade-off: stronger backbones often enhance Presence Stability (e.g., +3.4\% for MapTR \citep{liao2022maptr}) but can reduce Localization Stability (-12.8\% for MapTR \citep{liao2022maptr}), suggesting a potential focus on semantic over geometric consistency. 
% Critically, the model architecture dictates the backbone's importance. PivotNet's \citep{ding2023pivotnet} stability remains consistently high (mAS $\approx$ 84.5) across backbones, while GeMap \citep{zhang2024gemap} benefits significantly from Swin-T \citep{liu2021swintransformer}, achieving balanced gains in both mAP (+14.9\%) and mAS (+4.6\%).

\paragraph{Impact of Training Regimen.} Our analysis reveals distinct patterns in how extended training affects model performance across different architectures. As shown in Table \ref{tab:ablation_epoch}, while longer training epochs consistently improve accuracy (mAP increases ranging from +4.3\% to +40.1\%), the effects on temporal stability vary significantly. We observe three distinct learning behaviors. First, models like MapTR \citep{liao2022maptr} exhibit \emph{stability erosion}, where accuracy gains (+22.8\% for MapTR-50) come with stability degradation (-4.7\% mAS). Second, architectures such as MapQR \citep{liu2024mapqr} and PivotNet \citep{ding2023pivotnet} demonstrate \emph{stability saturation}, maintaining or slightly improving mAS (+3.2\% and +0.5\% respectively) while achieving accuracy improvements. Third, complex temporal models like MapTracker \citep{chen2024maptracker} show \emph{optimization sensitivity}, where extended training improves mAP (+3.7\% to +4.2\%) but leads to slight mAS reductions (-1.0\% to -1.4\%). These patterns underscore that temporal stability responds differently to extended training based on architectural inductive biases, suggesting that stability should be explicitly optimized rather than expected to emerge from accuracy-focused training alone.

\subsection{General Discussion}

Our benchmark reveals that \textbf{temporal stability (mAS) is an independent performance dimension} from accuracy (mAP), challenging the prevailing focus on single-frame precision. Models with high mAP can exhibit significant instability, underscoring the need for dual optimization. \textbf{Architectural choices induce distinct stability profiles.} Multi-sensor fusion improves accuracy but affects stability model-dependently. BEV encoders specialize differently: GKT \citep{chen2022GKT} favors detection consistency while BEVFormer \citep{li2024bevformer} variants reduce geometric jitter. Temporal fusion effectiveness hinges on architectural compatibility, with native designs outperforming retrofitted modules. \textbf{Training dynamics diverge by architecture.} Extended training improves accuracy consistently but affects stability variably, revealing three patterns: erosion (MapTR \citep{liao2022maptr}), saturation (MapQR \citep{liu2024mapqr}), and sensitivity (MapTracker \citep{chen2024maptracker}). This indicates stability requires explicit optimization rather than emerging implicitly from accuracy-focused training.

These findings advocate for \textbf{co-equal treatment of stability and accuracy} in evaluation and design. The substantial stability gaps among models (mAS: 66.6–91.9) highlight critical improvement opportunities. Future work should develop architectures that explicitly joint-optimize both criteria for trustworthy autonomous driving systems.

\section{Conclusion}
\label{sec:conclusion}
\vspace{-3mm}
In this work, we address the critical yet overlooked aspect of temporal stability in online HD mapping evaluation. While significant progress has been made in single-frame accuracy, the consistency of model outputs across sequential frames, which is essential for safe deployment, has remained largely unquantified. To bridge this gap, we introduce a multi-dimensional stability evaluation framework with novel metrics for presence, localization, and shape stability, integrated into a unified mean Average Stability (mAS) score. Extensive benchmarking demonstrates that accuracy (mAP) and stability (mAS) represent independent performance dimensions, challenging the assumption that accuracy optimization alone ensures real-world reliability. Our analysis further reveals how architectural choices, including temporal fusion strategies, sensor modality, training regimens, backbone designs, and BEV encoders, distinctly influence both accuracy and stability. By establishing the first stability-centric benchmark, we aim to shift community focus beyond accuracy alone and inspire the development of next-generation online HD mapping systems that achieve both high accuracy and temporal consistency, thereby advancing more robust and trustworthy autonomous driving.

\bibliography{iclr2026_conference}

\begin{thebibliography}{45}
\providecommand{\natexlab}[1]{#1}
\providecommand{\url}[1]{\texttt{#1}}
\expandafter\ifx\csname urlstyle\endcsname\relax
  \providecommand{\doi}[1]{doi: #1}\else
  \providecommand{\doi}{doi: \begingroup \urlstyle{rm}\Url}\fi

\bibitem[Caesar et~al.(2020)Caesar, Bankiti, Lang, Vora, Liong, Xu, Krishnan, Pan, Baldan, and Beijbom]{caesar2020nuscenes}
Holger Caesar, Varun Bankiti, Alex~H Lang, Sourabh Vora, Venice~Erin Liong, Qiang Xu, Anush Krishnan, Yu~Pan, Giancarlo Baldan, and Oscar Beijbom.
\newblock nuscenes: A multimodal dataset for autonomous driving.
\newblock In \emph{Proceedings of the IEEE/CVF conference on computer vision and pattern recognition}, pp.\  11621--11631, 2020.

\bibitem[Chen et~al.(2024)Chen, Wu, Tan, Ma, and Furukawa]{chen2024maptracker}
Jiacheng Chen, Yuefan Wu, Jiaqi Tan, Hang Ma, and Yasutaka Furukawa.
\newblock Maptracker: Tracking with strided memory fusion for consistent vector hd mapping.
\newblock In \emph{European Conference on Computer Vision}, pp.\  90--107. Springer, 2024.

\bibitem[Chen et~al.(2022)Chen, Cheng, Wang, Meng, Zhang, and Liu]{chen2022GKT}
Shaoyu Chen, Tianheng Cheng, Xinggang Wang, Wenming Meng, Qian Zhang, and Wenyu Liu.
\newblock Efficient and robust 2d-to-bev representation learning via geometry-guided kernel transformer.
\newblock \emph{arXiv preprint arXiv:2206.04584}, 2022.

\bibitem[Ding et~al.(2023)Ding, Qiao, Qiu, and Zhang]{ding2023pivotnet}
Wenjie Ding, Limeng Qiao, Xi~Qiu, and Chi Zhang.
\newblock Pivotnet: Vectorized pivot learning for end-to-end hd map construction.
\newblock In \emph{Proceedings of the IEEE/CVF International Conference on Computer Vision}, pp.\  3672--3682, 2023.

\bibitem[Dong et~al.(2023)Dong, Kang, Zhang, Zhu, Wang, Yang, Su, Wei, and Zhu]{dong2023robo3d1}
Yinpeng Dong, Caixin Kang, Jinlai Zhang, Zijian Zhu, Yikai Wang, Xiao Yang, Hang Su, Xingxing Wei, and Jun Zhu.
\newblock Benchmarking robustness of 3d object detection to common corruptions.
\newblock In \emph{Proceedings of the IEEE/CVF Conference on Computer Vision and Pattern Recognition}, pp.\  1022--1032, 2023.

\bibitem[Gu et~al.(2024)Gu, Song, Gilitschenski, Pavone, and Ivanovic]{gu2024traj}
Xunjiang Gu, Guanyu Song, Igor Gilitschenski, Marco Pavone, and Boris Ivanovic.
\newblock Producing and leveraging online map uncertainty in trajectory prediction.
\newblock In \emph{Proceedings of the IEEE/CVF Conference on Computer Vision and Pattern Recognition}, pp.\  14521--14530, 2024.

\bibitem[Hao et~al.(2024{\natexlab{a}})Hao, Li, Zhang, Li, Yin, Jung, Park, Yoo, Zhao, and Zhang]{hao2024mapdistill}
Xiaoshuai Hao, Ruikai Li, Hui Zhang, Dingzhe Li, Rong Yin, Sangil Jung, Seung-In Park, ByungIn Yoo, Haimei Zhao, and Jing Zhang.
\newblock Mapdistill: Boosting efficient camera-based hd map construction via camera-lidar fusion model distillation.
\newblock In \emph{European Conference on Computer Vision}, pp.\  166--183. Springer, 2024{\natexlab{a}}.

\bibitem[Hao et~al.(2024{\natexlab{b}})Hao, Wei, Yang, Zhao, Zhang, Zhou, Wang, Li, Kong, and Zhang]{hao2024your}
Xiaoshuai Hao, Mengchuan Wei, Yifan Yang, Haimei Zhao, Hui Zhang, Yi~Zhou, Qiang Wang, Weiming Li, Lingdong Kong, and Jing Zhang.
\newblock Is your hd map constructor reliable under sensor corruptions?
\newblock \emph{Advances in Neural Information Processing Systems}, 37:\penalty0 22441--22482, 2024{\natexlab{b}}.

\bibitem[Hao et~al.(2025{\natexlab{a}})Hao, Kong, Yin, Wang, Zhang, Diao, and Zhao]{hao2025safemap}
Xiaoshuai Hao, Lingdong Kong, Rong Yin, Pengwei Wang, Jing Zhang, Yunfeng Diao, and Shu Zhao.
\newblock Safemap: Robust hd map construction from incomplete observations.
\newblock \emph{arXiv preprint arXiv:2507.00861}, 2025{\natexlab{a}}.

\bibitem[Hao et~al.(2025{\natexlab{b}})Hao, Liu, Zhao, Ji, Wei, Zhao, Kong, Yin, and Liu]{hao2025msc}
Xiaoshuai Hao, Guanqun Liu, Yuting Zhao, Yuheng Ji, Mengchuan Wei, Haimei Zhao, Lingdong Kong, Rong Yin, and Yu~Liu.
\newblock Msc-bench: Benchmarking and analyzing multi-sensor corruption for driving perception.
\newblock \emph{arXiv preprint arXiv:2501.01037}, 2025{\natexlab{b}}.

\bibitem[He et~al.(2016)He, Zhang, Ren, and Sun]{he2016resnet}
Kaiming He, Xiangyu Zhang, Shaoqing Ren, and Jian Sun.
\newblock Deep residual learning for image recognition.
\newblock In \emph{Proceedings of the IEEE conference on computer vision and pattern recognition}, pp.\  770--778, 2016.

\bibitem[Hong et~al.(2022)Hong, Zhou, Zhu, Li, and Liu]{hong2022roboseg}
Fangzhou Hong, Hui Zhou, Xinge Zhu, Hongsheng Li, and Ziwei Liu.
\newblock Lidar-based 4d panoptic segmentation via dynamic shifting network.
\newblock \emph{arXiv preprint arXiv:2203.07186}, 2022.

\bibitem[Hu et~al.(2023)Hu, Yang, Chen, Li, Sima, Zhu, Chai, Du, Lin, Wang, et~al.]{hu2023uniad}
Yihan Hu, Jiazhi Yang, Li~Chen, Keyu Li, Chonghao Sima, Xizhou Zhu, Siqi Chai, Senyao Du, Tianwei Lin, Wenhai Wang, et~al.
\newblock Planning-oriented autonomous driving.
\newblock In \emph{Proceedings of the IEEE/CVF conference on computer vision and pattern recognition}, pp.\  17853--17862, 2023.

\bibitem[Jiang et~al.(2023)Jiang, Chen, Xu, Liao, Chen, Zhou, Zhang, Liu, Huang, and Wang]{jiang2023vad}
Bo~Jiang, Shaoyu Chen, Qing Xu, Bencheng Liao, Jiajie Chen, Helong Zhou, Qian Zhang, Wenyu Liu, Chang Huang, and Xinggang Wang.
\newblock Vad: Vectorized scene representation for efficient autonomous driving.
\newblock In \emph{Proceedings of the IEEE/CVF International Conference on Computer Vision}, pp.\  8340--8350, 2023.

\bibitem[Kim et~al.(2025)Kim, Woo, Heo, and Kim]{kim2025bridgeta}
Beomjun Kim, Suhan Woo, Sejong Heo, and Euntai Kim.
\newblock Bridgeta: Bridging the representation gap in knowledge distillation via teacher assistant for bird's eye view map segmentation.
\newblock \emph{arXiv preprint arXiv:2508.09599}, 2025.

\bibitem[Kong et~al.(2023)Kong, Niu, Xie, Hu, Ng, Cottereau, Zhang, Wang, Ooi, Zhu, et~al.]{kong2023robodepth}
Lingdong Kong, Yaru Niu, Shaoyuan Xie, Hanjiang Hu, Lai~Xing Ng, Benoit~R Cottereau, Liangjun Zhang, Hesheng Wang, Wei~Tsang Ooi, Ruijie Zhu, et~al.
\newblock The robodepth challenge: Methods and advancements towards robust depth estimation.
\newblock \emph{arXiv preprint arXiv:2307.15061}, 2023.

\bibitem[Kong et~al.(2025)Kong, Yang, Mei, Liu, Liang, Zhu, Lu, Yin, Hu, Jia, et~al.]{kong2025WM}
Lingdong Kong, Wesley Yang, Jianbiao Mei, Youquan Liu, Ao~Liang, Dekai Zhu, Dongyue Lu, Wei Yin, Xiaotao Hu, Mingkai Jia, et~al.
\newblock 3d and 4d world modeling: A survey.
\newblock \emph{arXiv preprint arXiv:2509.07996}, 2025.

\bibitem[Lee et~al.(2019)Lee, Hwang, Lee, Bae, and Park]{lee2019v299}
Youngwan Lee, Joong-won Hwang, Sangrok Lee, Yuseok Bae, and Jongyoul Park.
\newblock An energy and gpu-computation efficient backbone network for real-time object detection.
\newblock In \emph{Proceedings of the IEEE/CVF conference on computer vision and pattern recognition workshops}, pp.\  0--0, 2019.

\bibitem[Li et~al.(2022)Li, Wang, Wang, and Zhao]{li2022hdmapnet}
Qi~Li, Yue Wang, Yilun Wang, and Hang Zhao.
\newblock Hdmapnet: An online hd map construction and evaluation framework.
\newblock In \emph{2022 International Conference on Robotics and Automation (ICRA)}, pp.\  4628--4634. IEEE, 2022.

\bibitem[Li et~al.(2025)Li, Zhang, Qiu, Li, Liu, Wang, Li, Zhu, Gao, Lin, et~al.]{li2025onestage}
Yang Li, Zongzheng Zhang, Xuchong Qiu, Xinrun Li, Ziming Liu, Leichen Wang, Ruikai Li, Zhenxin Zhu, Huan-ang Gao, Xiaojian Lin, et~al.
\newblock Reusing attention for one-stage lane topology understanding.
\newblock \emph{arXiv preprint arXiv:2507.17617}, 2025.

\bibitem[Li et~al.(2024)Li, Wang, Li, Xie, Sima, Lu, Yu, and Dai]{li2024bevformer}
Zhiqi Li, Wenhai Wang, Hongyang Li, Enze Xie, Chonghao Sima, Tong Lu, Qiao Yu, and Jifeng Dai.
\newblock Bevformer: learning bird's-eye-view representation from lidar-camera via spatiotemporal transformers.
\newblock \emph{IEEE Transactions on Pattern Analysis and Machine Intelligence}, 2024.

\bibitem[Liao et~al.(2022)Liao, Chen, Wang, Cheng, Zhang, Liu, and Huang]{liao2022maptr}
Bencheng Liao, Shaoyu Chen, Xinggang Wang, Tianheng Cheng, Qian Zhang, Wenyu Liu, and Chang Huang.
\newblock Maptr: Structured modeling and learning for online vectorized hd map construction.
\newblock \emph{arXiv preprint arXiv:2208.14437}, 2022.

\bibitem[Liao et~al.(2025{\natexlab{a}})Liao, Chen, Yin, Jiang, Wang, Yan, Zhang, Li, Zhang, Zhang, et~al.]{liao2025diffusiondrive}
Bencheng Liao, Shaoyu Chen, Haoran Yin, Bo~Jiang, Cheng Wang, Sixu Yan, Xinbang Zhang, Xiangyu Li, Ying Zhang, Qian Zhang, et~al.
\newblock Diffusiondrive: Truncated diffusion model for end-to-end autonomous driving.
\newblock In \emph{Proceedings of the Computer Vision and Pattern Recognition Conference}, pp.\  12037--12047, 2025{\natexlab{a}}.

\bibitem[Liao et~al.(2025{\natexlab{b}})Liao, Chen, Zhang, Jiang, Zhang, Liu, Huang, and Wang]{liao2025maptrv2}
Bencheng Liao, Shaoyu Chen, Yunchi Zhang, Bo~Jiang, Qian Zhang, Wenyu Liu, Chang Huang, and Xinggang Wang.
\newblock Maptrv2: An end-to-end framework for online vectorized hd map construction.
\newblock \emph{International Journal of Computer Vision}, 133\penalty0 (3):\penalty0 1352--1374, 2025{\natexlab{b}}.

\bibitem[Lilja et~al.(2024)Lilja, Fu, Stenborg, and Hammarstrand]{lilja2024localization}
Adam Lilja, Junsheng Fu, Erik Stenborg, and Lars Hammarstrand.
\newblock Localization is all you evaluate: Data leakage in online mapping datasets and how to fix it.
\newblock In \emph{Proceedings of the IEEE/CVF Conference on Computer Vision and Pattern Recognition}, pp.\  22150--22159, 2024.

\bibitem[Liu et~al.(2024{\natexlab{a}})Liu, Wang, Li, Yang, Chen, and Zhu]{liu2024mgmap}
Xiaolu Liu, Song Wang, Wentong Li, Ruizi Yang, Junbo Chen, and Jianke Zhu.
\newblock Mgmap: Mask-guided learning for online vectorized hd map construction.
\newblock In \emph{Proceedings of the IEEE/CVF Conference on Computer Vision and Pattern Recognition}, pp.\  14812--14821, 2024{\natexlab{a}}.

\bibitem[Liu et~al.(2023)Liu, Yuan, Wang, Wang, and Zhao]{liu2023vectormapnet}
Yicheng Liu, Tianyuan Yuan, Yue Wang, Yilun Wang, and Hang Zhao.
\newblock Vectormapnet: End-to-end vectorized hd map learning.
\newblock In \emph{International Conference on Machine Learning}, pp.\  22352--22369. PMLR, 2023.

\bibitem[Liu et~al.(2021)Liu, Lin, Cao, Hu, Wei, Zhang, Lin, and Guo]{liu2021swintransformer}
Ze~Liu, Yutong Lin, Yue Cao, Han Hu, Yixuan Wei, Zheng Zhang, Stephen Lin, and Baining Guo.
\newblock Swin transformer: Hierarchical vision transformer using shifted windows.
\newblock In \emph{Proceedings of the IEEE/CVF international conference on computer vision}, pp.\  10012--10022, 2021.

\bibitem[Liu et~al.(2022)Liu, Tang, Amini, Yang, Mao, Rus, and Han]{liu2022bevfusion}
Zhijian Liu, Haotian Tang, Alexander Amini, Xinyu Yang, Huizi Mao, Daniela Rus, and Song Han.
\newblock Bevfusion: Multi-task multi-sensor fusion with unified bird's-eye view representation.
\newblock \emph{arXiv preprint arXiv:2205.13542}, 2022.

\bibitem[Liu et~al.(2024{\natexlab{b}})Liu, Zhang, Liu, Zhao, and Xu]{liu2024mapqr}
Zihao Liu, Xiaoyu Zhang, Guangwei Liu, Ji~Zhao, and Ningyi Xu.
\newblock Leveraging enhanced queries of point sets for vectorized map construction.
\newblock In \emph{European Conference on Computer Vision}, pp.\  461--477. Springer, 2024{\natexlab{b}}.

\bibitem[Paek et~al.(2022)Paek, Kong, and Wijaya]{paek2022K-Radar}
Dong-Hee Paek, Seung-Hyun Kong, and Kevin~Tirta Wijaya.
\newblock K-radar: 4d radar object detection for autonomous driving in various weather conditions.
\newblock \emph{Advances in Neural Information Processing Systems}, 35:\penalty0 3819--3829, 2022.

\bibitem[Qiao et~al.(2023)Qiao, Ding, Qiu, and Zhang]{qiao2023bemapnet}
Limeng Qiao, Wenjie Ding, Xi~Qiu, and Chi Zhang.
\newblock End-to-end vectorized hd-map construction with piecewise bezier curve.
\newblock In \emph{Proceedings of the IEEE/CVF Conference on Computer Vision and Pattern Recognition}, pp.\  13218--13228, 2023.

\bibitem[Tan \& Le(2019)Tan and Le]{tan2019efficientnet}
Mingxing Tan and Quoc Le.
\newblock Efficientnet: Rethinking model scaling for convolutional neural networks.
\newblock 2019.

\bibitem[Wang et~al.(2020)Wang, Sun, Kortylewski, and Yuille]{wang2020robo2D}
Angtian Wang, Yihong Sun, Adam Kortylewski, and Alan~L Yuille.
\newblock Robust object detection under occlusion with context-aware compositionalnets.
\newblock In \emph{Proceedings of the IEEE/CVF conference on computer vision and pattern recognition}, pp.\  12645--12654, 2020.

\bibitem[Wang et~al.(2023)Wang, Li, Liu, Liu, and Zhu]{wang2023lidar2map}
Song Wang, Wentong Li, Wenyu Liu, Xiaolu Liu, and Jianke Zhu.
\newblock Lidar2map: In defense of lidar-based semantic map construction using online camera distillation.
\newblock In \emph{Proceedings of the IEEE/CVF Conference on Computer Vision and Pattern Recognition}, pp.\  5186--5195, 2023.

\bibitem[Xie et~al.(2023)Xie, Kong, Zhang, Ren, Pan, Chen, and Liu]{xie2023robobev}
Shaoyuan Xie, Lingdong Kong, Wenwei Zhang, Jiawei Ren, Liang Pan, Kai Chen, and Ziwei Liu.
\newblock Robobev: Towards robust bird's eye view perception under corruptions.
\newblock \emph{arXiv preprint arXiv:2304.06719}, 2023.

\bibitem[Xie et~al.(2025)Xie, Kong, Zhang, Ren, Pan, Chen, and Liu]{xie2025robobevv2}
Shaoyuan Xie, Lingdong Kong, Wenwei Zhang, Jiawei Ren, Liang Pan, Kai Chen, and Ziwei Liu.
\newblock Benchmarking and improving bird's eye view perception robustness in autonomous driving.
\newblock \emph{IEEE Transactions on Pattern Analysis and Machine Intelligence}, 2025.

\bibitem[Yan et~al.(2018)Yan, Mao, and Li]{yan2018second}
Yan Yan, Yuxing Mao, and Bo~Li.
\newblock Second: Sparsely embedded convolutional detection.
\newblock \emph{Sensors}, 18\penalty0 (10):\penalty0 3337, 2018.

\bibitem[Yan et~al.(2025)Yan, Li, Cui, Li, Jiang, Ren, Li, Li, Wen, and Yu]{yan2025mapkd}
Ziyang Yan, Ruikai Li, Zhiyong Cui, Bohan Li, Han Jiang, Yilong Ren, Aoyong Li, Zhenning Li, Sijia Wen, and Haiyang Yu.
\newblock Mapkd: Unlocking prior knowledge with cross-modal distillation for efficient online hd map construction.
\newblock \emph{arXiv preprint arXiv:2508.15653}, 2025.

\bibitem[Yuan et~al.(2024)Yuan, Liu, Wang, Wang, and Zhao]{yuan2024streammapnet}
Tianyuan Yuan, Yicheng Liu, Yue Wang, Yilun Wang, and Hang Zhao.
\newblock Streammapnet: Streaming mapping network for vectorized online hd map construction.
\newblock In \emph{Proceedings of the IEEE/CVF Winter Conference on Applications of Computer Vision}, pp.\  7356--7365, 2024.

\bibitem[Zhang et~al.(2023)Zhang, Lin, Wu, Luo, Xue, Lu, Wang, et~al.]{zhang2023mapvr}
Gongjie Zhang, Jiahao Lin, Shuang Wu, Zhipeng Luo, Yang Xue, Shijian Lu, Zuoguan Wang, et~al.
\newblock Online map vectorization for autonomous driving: A rasterization perspective.
\newblock \emph{Advances in Neural Information Processing Systems}, 36:\penalty0 31865--31877, 2023.

\bibitem[Zhang et~al.(2024{\natexlab{a}})Zhang, Liu, Liu, Xu, Liu, and Zhao]{zhang2024hrmapnet}
Xiaoyu Zhang, Guangwei Liu, Zihao Liu, Ningyi Xu, Yunhui Liu, and Ji~Zhao.
\newblock Enhancing vectorized map perception with historical rasterized maps.
\newblock In \emph{European Conference on Computer Vision}, pp.\  422--439. Springer, 2024{\natexlab{a}}.

\bibitem[Zhang et~al.(2024{\natexlab{b}})Zhang, Zhang, Ding, Jin, and Yue]{zhang2024gemap}
Zhixin Zhang, Yiyuan Zhang, Xiaohan Ding, Fusheng Jin, and Xiangyu Yue.
\newblock Online vectorized hd map construction using geometry.
\newblock In \emph{European Conference on Computer Vision}, pp.\  73--90. Springer, 2024{\natexlab{b}}.

\bibitem[Zhang et~al.(2025)Zhang, Qiu, Zhang, Zheng, Gu, Chi, Gao, Wang, Liu, Li, et~al.]{zhang2025irostraj}
Zongzheng Zhang, Xuchong Qiu, Boran Zhang, Guantian Zheng, Xunjiang Gu, Guoxuan Chi, Huan-ang Gao, Leichen Wang, Ziming Liu, Xinrun Li, et~al.
\newblock Delving into mapping uncertainty for mapless trajectory prediction.
\newblock \emph{arXiv preprint arXiv:2507.18498}, 2025.

\bibitem[Zhu et~al.(2023)Zhu, Zhang, Chen, Dong, Zhao, Ding, Zhong, and Zheng]{zhu2023robo3d2}
Zijian Zhu, Yichi Zhang, Hai Chen, Yinpeng Dong, Shu Zhao, Wenbo Ding, Jiachen Zhong, and Shibao Zheng.
\newblock Understanding the robustness of 3d object detection with bird's-eye-view representations in autonomous driving.
\newblock In \emph{Proceedings of the IEEE/CVF Conference on Computer Vision and Pattern Recognition}, pp.\  21600--21610, 2023.

\end{thebibliography}
\bibliographystyle{iclr2026_conference}

\clearpage

\appendix
{\LARGE  Appendix}

In the appendix, we supply further details on the proposed stability evaluation framework, the benchmark setup, experimental analyses, and visualizations that are omitted from the main paper for brevity. The appendix is structured as follows:

\begin{itemize}
    \item Sec.~\ref{sec:appendix_3_LLMS} provides additional statements on the use of Large Language Models (LLMs) in this work.
    \item Sec.~\ref{sec:appendix_4_algorithm} presents implementation details of the stability evaluation algorithm pipeline.
    \item Sec.~\ref{sec:appendix_5_exp_setting} offers supplementary experimental setups and related ablation studies.
    \item Sec.~\ref{sec:appendix_6_model_details} presents additional analyses from 10 different models.
    \item Sec.~\ref{sec:appendix_7_vis} displays supplementary visualizations, encompassing additional mAP \textit{vs.} mAS comparisons , and illustrations of how unstable predictions affect downstream tasks.
    %\item Sec.\ref{sec:appendix_8_discussion} concludes our experiments and provides a comprehensive discussion of the findings.
    \item Sec.~\ref{sec:appendix_9_limit_future} discusses the limitations of our work and provides an outlook on future work.
    
\end{itemize}

\section{Statement on the Use of Large Language Models (LLMs)}
\label{sec:appendix_3_LLMS}
In the preparation of this paper, large language models (LLMs) were used solely as an assistive tool for writing refinement and polishing. The core research ideas, theoretical framework, experimental design, data analysis, and result interpretation were entirely conceived and conducted by the human authors. The LLM was employed after the intellectual substance of the work was fully established, specifically to assist with improving grammatical correctness, sentence fluency, and overall clarity of expression in certain parts of the manuscript. It did not contribute to the scientific ideation, methodological development, or conclusions of the research. All final content was thoroughly reviewed, verified, and approved by the authors.

\section{Stability Evaluation Algorithm with Additional Details}
\label{sec:appendix_4_algorithm}
This section provides comprehensive algorithmic details for the multi-dimensional map stability evaluation framework introduced in Section~\ref{sec:eval_framework}. The complete pipeline consists of four main stages: temporal sampling, cross-frame instance matching, geometric alignment and resampling, and stability metric computation. Each stage is implemented through carefully designed algorithms that ensure robust and reproducible evaluation of temporal stability in online HD mapping.

% \subsection{Data Preprocessing}
% Algorithm~\ref{alg:stability_preprocessing} presents the complete preprocessing pipeline that orchestrates the entire stability evaluation process. 
\subsection{Temporal Sampling}

Algorithm~\ref{alg:temporal_sampling} implements the temporal sampling stage that constructs frame pairs for stability analysis. The algorithm randomly selects subsequent frames within a predefined maximum temporal interval $M$ for each anchor frame, ensuring comprehensive coverage of temporal variations while maintaining computational efficiency. This approach generates a sampling set $S$ of size $|S| = L - M$, providing the foundational inputs for subsequent stability analysis.

% Temporal Sampling Algorithm

\begin{algorithm}[htbp]
\caption{Temporal Sampling Algorithm}
\label{alg:temporal_sampling}
\begin{algorithmic}[1]
\REQUIRE Model output frame sequence $\{D_1, D_2, \ldots, D_L\}$, maximum temporal interval $M$
\ENSURE Frame pair sampling set $S$

\STATE \textbf{// Stage 1: Temporal Sampling}
\STATE $S \leftarrow \emptyset$ \COMMENT{Initialize sampling set}
\FOR{$t = 1$ to $L - M$}
    \STATE $k \leftarrow \text{RandomSample}(1, M)$ \COMMENT{Random sampling within $[1, M]$ range}
    \STATE $S \leftarrow S \cup \{(D_t, D_{t+k})\}$ \COMMENT{Add frame pair to sampling set}
\ENDFOR

\RETURN $S$
\end{algorithmic}
\end{algorithm}

\subsection{Cross-Frame Instance Matching}

Algorithm~\ref{alg:cross_frame_matching} implements the cross-frame instance matching stage that establishes correspondence between map elements across temporal frames. 

% Cross-Frame Instance Matching Algorithm

\begin{algorithm}[htbp]
\caption{Cross-Frame Instance Matching Algorithm}
\label{alg:cross_frame_matching}
\begin{algorithmic}[1]
\REQUIRE Frame pair sampling set $S$
\ENSURE Matched polyline pairs $\mathcal{M}$

\STATE \textbf{// Stage 2: Cross-Frame Instance Matching}
\STATE $\mathcal{M} \leftarrow \emptyset$ \COMMENT{Initialize matching result set}
\FOR{each $(D_t, D_{t+k}) \in S$}
    \STATE \textbf{// Step 2.1: Frame-to-GT Matching}
    \STATE $\text{matches}_t \leftarrow \text{HungarianMatching}(D_t, \text{GT}_t)$
    \STATE $\text{matches}_{t+k} \leftarrow \text{HungarianMatching}(D_{t+k}, \text{GT}_{t+k})$
    
    \STATE \textbf{// Step 2.2: GT-based Association}
    \STATE $E \leftarrow \text{FindCommonGTInstances}(\text{matches}_t, \text{matches}_{t+k})$
    \FOR{each $e \in E$}
        \STATE $\text{poly}_t(e) \leftarrow \text{GetPolyline}(\text{matches}_t, e)$
        \STATE $\text{poly}_{t+k}(e) \leftarrow \text{GetPolyline}(\text{matches}_{t+k}, e)$
        \STATE $\mathcal{M} \leftarrow \mathcal{M} \cup \{(\text{poly}_{t+k}(e), \text{poly}_t(e))\}$
    \ENDFOR
\ENDFOR

\RETURN $\mathcal{M}$
\end{algorithmic}
\end{algorithm}

The algorithm employs a two-step strategy: first matching predictions to ground truth within each frame using the algorithm \ref{alg:hungarian_matching}, then establishing temporal correspondence through ground truth-based association.

% Hungarian Matching Sub-algorithm

\begin{algorithm}[htbp]
\caption{Hungarian Matching Sub-algorithm}
\label{alg:hungarian_matching}
\begin{algorithmic}[1]
\REQUIRE Prediction frame $D$, ground truth frame $\text{GT}$, cost function $C$
\ENSURE Matching result $\text{matches}$

\STATE $P \leftarrow \text{GetPolygons}(D)$ \COMMENT{Get prediction polygons}
\STATE $G \leftarrow \text{GetPolygons}(\text{GT})$ \COMMENT{Get ground truth polygons}
\STATE $n \leftarrow |P|$, $m \leftarrow |G|$

\STATE \textbf{// Build cost matrix}
\STATE $C_{\text{matrix}} \leftarrow \text{zeros}(n \times m)$
\FOR{$i = 1$ to $n$}
    \FOR{$j = 1$ to $m$}
        \STATE $C_{\text{matrix}}[i,j] \leftarrow C(P[i], G[j])$ \COMMENT{Geometric and semantic similarity cost}
    \ENDFOR
\ENDFOR

\STATE \textbf{// Execute Hungarian algorithm}
\STATE $\text{matches} \leftarrow \text{HungarianAlgorithm}(C_{\text{matrix}})$
\RETURN $\text{matches}$
\end{algorithmic}
\end{algorithm}

 This indirect matching approach leverages the consistency of ground truth annotations to overcome the inherent inconsistencies in model predictions, yielding a set of matched instance pairs for each frame pair.

\subsection{Geometric Alignment and Resampling}

Algorithm~\ref{alg:geometric_alignment} implements the geometric alignment and resampling stage that ensures spatially consistent comparison between matched polylines. 

% Geometric Alignment and Resampling Algorithm

\begin{algorithm}[htbp]
\caption{Geometric Alignment and Resampling Algorithm}
\label{alg:geometric_alignment}
\begin{algorithmic}[1]
\REQUIRE Matched polyline pairs $\mathcal{M}$, perception range $[x_{\min}, x_{\max}, y_{\min}, y_{\max}]$, resampling points $N$
\ENSURE Matched and aligned polyline pairs $\mathcal{M}^{\text{sample}}$

\STATE \textbf{// Stage 3: Geometric Alignment and Resampling}
\STATE $\mathcal{M}^{\text{sample}} \leftarrow \emptyset$ \COMMENT{Initialize sampled polyline pairs set}
\FOR{each $(\text{poly}_{t+k}(e), \text{poly}_t(e)) \in \mathcal{M}$}
    \STATE \textbf{// Step 3.1: Coordinate Transformation}
    \STATE $\text{poly}_{t \to t+k}(e) \leftarrow T_{\text{world} \to t+k} \cdot T_{t \to \text{world}} \cdot \text{poly}_t(e)$
    
    \STATE \textbf{// Step 3.2: Perception Range Filtering}
    \STATE $\text{poly}_{t \to t+k}(e) \leftarrow \text{ClipToPerceptionRange}(\text{poly}_{t \to t+k}(e), [x_{\min}, x_{\max}, y_{\min}, y_{\max}])$
    
    \STATE \textbf{// Step 3.3: Uniform Resampling}
    \STATE $[x_{\min}^p, x_{\max}^p] \leftarrow \text{GetCommonXRange}(\text{poly}_{t+k}(e), \text{poly}_{t \to t+k}(e))$
    \FOR{$i = 1$ to $N$}
        \STATE $x_i \leftarrow x_{\min}^p + (i-1) \cdot \frac{x_{\max}^p - x_{\min}^p}{N-1}$
        \STATE $y_{t+k}(x_i) \leftarrow \text{InterpolateY}(\text{poly}_{t+k}(e), x_i)$
        \STATE $y_t(x_i) \leftarrow \text{InterpolateY}(\text{poly}_{t \to t+k}(e), x_i)$
    \ENDFOR
    \STATE $\text{poly}_{t+k}^{\text{sample}}(e) \leftarrow \{(x_i, y_{t+k}(x_i)) \mid i = 1, 2, \ldots, N\}$
    \STATE $\text{poly}_{t}^{\text{sample}}(e) \leftarrow \{(x_i, y_t(x_i)) \mid i = 1, 2, \ldots, N\}$
    \STATE $\mathcal{M}^{\text{sample}} \leftarrow \mathcal{M}^{\text{sample}} \cup \{(\text{poly}_{t+k}^{\text{sample}}(e), \text{poly}_{t}^{\text{sample}}(e), e)\}$
\ENDFOR

\RETURN $\mathcal{M}^{\text{sample}}$
\end{algorithmic}
\end{algorithm}

The algorithm transforms historical polylines into the current frame's coordinate system, applies algorithm \ref{alg:perception_range_filtering} to ensure evaluation consistency, and performs uniform resampling along the $x$-axis. This process guarantees spatially aligned and comparable point sets for subsequent stability analysis, returning a comprehensive set of matched and aligned polyline pairs, each annotated with the corresponding map element identifier.

% Perception Range Filtering Sub-algorithm

\begin{algorithm}[htbp]
\caption{Perception Range Filtering Sub-algorithm}
\label{alg:perception_range_filtering}
\begin{algorithmic}[1]
\REQUIRE Transformed polyline $\text{poly}_{t \to t+k}(e)$, perception range $[x_{\min}, x_{\max}, y_{\min}, y_{\max}]$
\ENSURE Filtered polyline $\text{poly}_{\text{filtered}}$

\STATE $\text{poly}_{\text{filtered}} \leftarrow \emptyset$
\FOR{each point $(x, y) \in \text{poly}_{t \to t+k}(e)$}
    \IF{$x_{\min} \leq x \leq x_{\max}$ AND $y_{\min} \leq y \leq y_{\max}$}
        \STATE $\text{poly}_{\text{filtered}} \leftarrow \text{poly}_{\text{filtered}} \cup \{(x, y)\}$
    \ENDIF
\ENDFOR
\RETURN $\text{poly}_{\text{filtered}}$
\end{algorithmic}
\end{algorithm}

\subsection{Stability Metric Computation}

Algorithm~\ref{alg:stability_metric_computation} implements the core stability metric computation that quantifies temporal stability across three dimensions: presence, localization, and shape. The algorithm processes each matched polyline pair to compute individual stability scores, then aggregates these scores at the class and model levels. The presence stability evaluates detection consistency, the localization stability quantifies positional jitter, and the shape stability assesses geometric consistency through curvature comparison. 

% Multi-dimensional Map Stability Evaluation Framework - Stability Metric Computation

\begin{algorithm}[htbp]
\caption{Multi-dimensional Map Stability Evaluation Framework - Stability Metric Computation}
\label{alg:stability_metric_computation}
\begin{algorithmic}[1]
\REQUIRE Matched and aligned polyline pairs $\mathcal{M}^{\text{sample}}$, detection threshold $\tau$, weighting parameter $\omega$, scaling parameter $\beta$
\ENSURE Overall model stability score mAS

\STATE \textbf{// Stage 4: Stability Metric Computation}
\STATE $\mathcal{C} \leftarrow \text{GetAllClasses}(\mathcal{M}^{\text{sample}})$ \COMMENT{Get all classes}
\FOR{each $\text{class} \in \mathcal{C}$}
    \STATE $\mathcal{I}_{\text{class}} \leftarrow \text{GetInstancesOfClass}(\mathcal{M}^{\text{sample}}, \text{class})$
    \STATE $\text{stability}_{\text{class}} \leftarrow 0$
    
    \FOR{each $(\text{poly}_{t+k}^{\text{sample}}(e), \text{poly}_{t}^{\text{sample}}(e), e) \in \mathcal{I}_{\text{class}}$}
        \STATE \textbf{// Compute Presence Stability}
        \IF{$\text{score}_{t+k}(e) \geq \tau$ AND $\text{score}_t(e) \geq \tau$ OR $\text{score}_{t+k}(e) < \tau$ AND $\text{score}_t(e) < \tau$}
            \STATE $\text{Presence}(e) \leftarrow 1$
        \ELSE
            \STATE $\text{Presence}(e) \leftarrow 0.5$ \COMMENT{Flickering case}
        \ENDIF
        
        \STATE \textbf{// Compute Localization Stability}
        \STATE $\text{avg\_deviation} \leftarrow \frac{1}{N} \sum_{i=1}^{N} |y_{t+k}(x_i) - y_t(x_i)|$
        \STATE $\text{Loc}(e) \leftarrow \beta \cdot \text{avg\_deviation}$
        
        \STATE \textbf{// Compute Shape Stability}
        \STATE $\kappa_{t+k} \leftarrow \text{ComputeCurvature}(\text{poly}_{t+k}^{\text{sample}}(e))$
        \STATE $\kappa_t \leftarrow \text{ComputeCurvature}(\text{poly}_{t}^{\text{sample}}(e))$
        \STATE $\text{Shape}(e) \leftarrow 1 - \frac{|\kappa_{t+k} - \kappa_t|}{\pi}$
        
        \STATE \textbf{// Compute Comprehensive Stability}
        \STATE $\text{Stability}(e) \leftarrow \text{Presence}(e) \cdot [\omega \cdot \text{Loc}(e) + (1-\omega) \cdot \text{Shape}(e)]$
        \STATE $\text{stability}_{\text{class}} \leftarrow \text{stability}_{\text{class}} + \text{Stability}(e)$
    \ENDFOR
    \STATE $\text{Stability}_{\text{class}} \leftarrow \frac{\text{stability}_{\text{class}}}{|\mathcal{I}_{\text{class}}|}$
\ENDFOR

\STATE \textbf{// Compute Overall Model Stability}
\STATE $\text{mAS} \leftarrow \frac{1}{|\mathcal{C}|} \sum_{\text{class} \in \mathcal{C}} \text{Stability}_{\text{class}}$
\RETURN $\text{mAS}$
\end{algorithmic}
\end{algorithm}

The specific implementation of the ComputeCurvature function mentioned in Algorithm \ref{alg:stability_metric_computation} is detailed in Algorithm \ref{alg:curvature_computation}, which approximates polyline curvature by computing the average angles between consecutive segments, thereby providing a robust geometric measurement method with translation and rotation invariance. 

% Curvature Computation Sub-algorithm

\begin{algorithm}[htbp]
\caption{Curvature Computation Sub-algorithm}
\label{alg:curvature_computation}
\begin{algorithmic}[1]
\REQUIRE Resampled polyline $\text{poly}^{\text{sample}} = \{(x_i, y_i) \mid i = 1, 2, \ldots, N\}$
\ENSURE Curvature $\kappa$

\STATE $\kappa \leftarrow 0$
\FOR{$j = 1$ to $N-1$}
    \STATE $\vec{v_j} \leftarrow (x_{j+1} - x_j, y_{j+1} - y_j)$ \COMMENT{Compute vector $\vec{v_j}$}
    \STATE $\vec{v_{j+1}} \leftarrow (x_{j+2} - x_{j+1}, y_{j+2} - y_{j+1})$ \COMMENT{Compute vector $\vec{v_{j+1}}$}
    \STATE $\theta_j \leftarrow \cos^{-1}\left(\frac{\vec{v_j} \cdot \vec{v_{j+1}}}{|\vec{v_j}| \cdot |\vec{v_{j+1}}|}\right)$ \COMMENT{Compute angle}
    \STATE $\kappa \leftarrow \kappa + \theta_j$
\ENDFOR
\STATE $\kappa \leftarrow \frac{\kappa}{N-1}$ \COMMENT{Compute average curvature}
\RETURN $\kappa$
\end{algorithmic}
\end{algorithm}

This curvature-based approach enables effective comparison of geometric consistency across temporal frames while capturing subtle shape variations that may indicate instability in the model's predictions.

Upon completion of all sub-metric evaluations, the final mean Average Stability (mAS) score is computed, providing a holistic measure of the model's temporal stability across all evaluated classes.
%\input{algorithms/1_HD Map Prediction-GT Matching}

% \clearpage

\section{Further Details on the Experimental Setup}
\label{sec:appendix_5_exp_setting}

\subsection{Supplemental Details on the Experimental Setup}

Prior to conducting experiments, it is necessary to configure certain hyperparameters. This section primarily elaborates on the detailed configurations of these hyperparameters adopted in our study, along with the rationale for these choices:

\begin{itemize}

\item Maximum frame interval (\(M\)=2): The configuration of different frame intervals essentially represents distinct evaluation scenarios for stability assessment, each carrying unique implications. Therefore, in addition to the experiments with \(M\)=2 presented in the main text, as shown in chapter \ref{subsec:different_M_experiment}, we have conducted supplementary experiments with \(M\)=3, \(M\)=5, and \(M\)=10 to provide as comprehensive a stability evaluation as possible for existing models.

\item Number of resampling points (\(N\)=100):  The purpose of resampling is to adjust the distribution of map points on two instance polylines to be identical, thereby facilitating the calculation of stability metrics. Thus, the value of N should not be set too small to avoid undue influence from individual outliers on the instances in subsequent computations. However, beyond a sufficient threshold, variations in N do not significantly affect stability evaluation outcomes. Conversely, excessively large values of N may substantially reduce computational efficiency. Balancing resampling granularity and computational cost, we ultimately set N = 100. This value can be appropriately adjusted in different experimental settings.

\item Position Stability Scaling Factor ($\beta$=15.0): The specific implication of this scaling factor is that when the distance between two matched map points (i.e., points sampled at identical x-values on matched map instances) in adjacent frames equals $\beta$, their positional stability is considered zero. Consequently, the value of $\beta$ corresponds to the distance threshold representing extreme instability. Typically, we define such extreme cases using the shorter map radius (half the length of the map's shorter side). In prevailing map construction paradigms, the standard map range is generally defined as $\text{x} \in [-15, 15], \text{y} \in [-30, 30]$. Therefore, in our experiments, $\beta$ is set to 15.0.

\end{itemize}

\begin{table}[htbp]
\centering
%\fontsize{8pt}{10pt}\selectfont

\caption{Ablation Study on the Temporal Sampling Interval (\(M\)=3)}
\fontsize{6pt}{8pt}\selectfont
\label{tab:appendix_intreval_1}
\begin{tabular}[t]{r|r|cccc|c|ccc|c}
\toprule
\rowcolor{mygray}
\bf Method & 
\bf Venue &
\bf Temp &
\bf Modal &
\bf BEV Encoder &
\bf Epoch &
\bf mAP$\uparrow$ &
\bf Presence$\uparrow$ &
\bf Loc$\uparrow$ &
\bf Shape$\uparrow$ &
\bf \bf mAS$\uparrow$ 
\\
\midrule
\rowcolor{new_blue!3} MapTR & ICLR'23 & \XSolidBrush & C & GKT & 24 & 44.1 & 89.3 & 59.9 &  90.6 & 67.6\\
\rowcolor{new_blue!3} MapTR & ICLR'23 & \XSolidBrush & C \& L & GKT & 24 &  62.8 & 91.0 & 69.3 & 91.8  & 73.7 \\
\rowcolor{new_blue!3} BeMapNet & CVPR'23 & \XSolidBrush & C & IPM-PE & 30 & 61.4 & 100.0 & 58.8 & 97.7 & 78.2\\
\rowcolor{new_blue!3} PivotNet & ICCV'23 & \XSolidBrush & C & PersFormer & 30 & 57.1 & 100.0 & 45.2 & 98.3 & 71.7\\
\rowcolor{new_blue!3} MapTRv2 & IJCV'24 & \XSolidBrush & C & BEVPool & 24 & 61.4 & 90.6 & 59.5 &  91.8 & 69.0\\
\rowcolor{new_blue!3} GeMap & ECCV'24 & \XSolidBrush & C & BEVFormer-1 & 24 & 51.3 & 90.9 & 66.6 & 92.9 & 73.3 \\
\rowcolor{new_blue!3} MGMap & CVPR'24 & \XSolidBrush & C & BEVFormer-1 & 24 & 57.9 & 91.8 & 68.8 & 92.3 & 74.4\\
\rowcolor{new_blue!3} MapQR & ECCV'24 & \XSolidBrush & C & BEVFormer-3 & 24 & 66.4 & 89.4 & 66.9 & 91.1 & 73.4\\
\midrule
\rowcolor{new_orange!5} MapTR & ICLR'23 & \Checkmark & C & GKT & 24 & 51.3 & 86.8 & 55.2 &  89.1 & 62.9 \\
\rowcolor{new_orange!5} StreamMapNet & WACV'24 & \Checkmark & C & BEVFormer-1 & 30 & 63.3 & 96.9 & 96.6 & 95.8 & 93.2\\
\rowcolor{new_orange!5} MapTracker & ECCV'24 & \Checkmark & C & BEVFormer-2 & 72 & 75.95 & 93.7 & 96.2  & 93.4 & 88.7\\
\rowcolor{new_orange!5} HRMapNet & ECCV'24 & \Checkmark & C & BEVFormer-1 & 24 & 67.2 & 91.2 & 70.4 & 92.2 & 75.4\\
\bottomrule
\end{tabular}
\end{table}

\begin{table}[htbp]
\centering
%\fontsize{8pt}{10pt}\selectfont

\caption{Ablation Study on the Temporal Sampling Interval (\(M\)=5)}
\fontsize{6pt}{8pt}\selectfont
\label{tab:appendix_intreval_2}
\begin{tabular}[t]{r|r|cccc|c|ccc|c}
\toprule
\rowcolor{mygray}
\bf Method & 
\bf Venue &
\bf Temp &
\bf Modal &
\bf BEV Encoder &
\bf Epoch &
\bf mAP$\uparrow$ &
\bf Presence$\uparrow$ &
\bf Loc$\uparrow$ &
\bf Shape$\uparrow$ &
\bf \bf mAS$\uparrow$ 
\\
\midrule
\rowcolor{new_blue!3} MapTR & ICLR'23 & \XSolidBrush & C & GKT & 24 & 44.1 & 89.0 & 64.7 &  90.0 & 68.8\\
 MapTR & ICLR'23 & \XSolidBrush & C \& L & GKT & 24 &  62.8 & 89.4 & 69.2 & 91.0  & 72.1 \\
\rowcolor{new_blue!3} BeMapNet & CVPR'23 & \XSolidBrush & C & IPM-PE & 30 & 61.4 & 100.0 & 50.0 & 97.5 & 73.7\\
 PivotNet & ICCV'23 & \XSolidBrush & C & PersFormer & 30 & 57.1 & 100.0 & 41.5 & 98.3 & 70.0\\
\rowcolor{new_blue!3} MapTRv2 & IJCV'24 & \XSolidBrush & C & BEVPool & 24 & 61.4 & 89.0 & 52.8 &  90.8 & 64.8\\
 GeMap & ECCV'24 & \XSolidBrush & C & BEVFormer-1 & 24 & 51.3 & 88.8 & 61.7 & 91.6 & 69.3 \\
\rowcolor{new_blue!3} MGMap & CVPR'24 & \XSolidBrush & C & BEVFormer-1 & 24 & 57.9 & 90.0 & 67.1 & 91.5 & 71.9\\
MapQR & ECCV'24 & \XSolidBrush & C & BEVFormer-3 & 24 & 66.4 & 88.1 & 59.9 & 89.6 & 67.4\\
\rowcolor{new_blue!5} MapTR & ICLR'23 & \Checkmark & C & GKT & 24 & 51.3 & 85.0 & 55.8 &  90.3 & 61.8 \\
%\rowcolor{new_orange!5} StreamMapNet & WACV'24 & \Checkmark & C & BEVFormer-1 & 30 & 63.3 & \todo{X} & \todo{X} & \todo{X} & \todo{X}\\
%\rowcolor{new_orange!5} MapTracker & ECCV'24 & \Checkmark & C & BEVFormer-2 & 72 & 75.95 & \todo{X} & \todo{X} & \todo{X} & \todo{X}\\
 HRMapNet & ECCV'24 & \Checkmark & C & BEVformer-1 & 24 & 67.2 & 89.6 & 70.0 & 92.1 & 73.8\\
\bottomrule
\end{tabular}
\end{table}

\begin{table}[htbp]
\centering
%\fontsize{8pt}{10pt}\selectfont

\caption{Ablation Study on the Temporal Sampling Interval (\(M\)=10)}
\fontsize{6pt}{8pt}\selectfont
\label{tab:appendix_intreval_3}
\begin{tabular}[t]{r|r|cccc|c|ccc|c}
\toprule
\rowcolor{mygray}
\bf Method & 
\bf Venue &
\bf Temp &
\bf Modal &
\bf BEV Encoder &
\bf Epoch &
\bf mAP$\uparrow$ &
\bf Presence$\uparrow$ &
\bf Loc$\uparrow$ &
\bf Shape$\uparrow$ &
\bf \bf mAS$\uparrow$ 
\\
\midrule
\rowcolor{new_blue!3} MapTR & ICLR'23 & \XSolidBrush & C & GKT & 24 & 44.1 & 88.0 & 47.4 &  90.1 & 61.9\\
MapTR & ICLR'23 & \XSolidBrush & C \& L & GKT & 24 &  62.8 & 89.0 & 58.2 & 90.9  & 66.5 \\
\rowcolor{new_blue!3} BeMapNet & CVPR'23 & \XSolidBrush & C & IPM-PE & 30 & 61.4 & 100.0 & 41.6 & 97.5 & 69.5\\
PivotNet & ICCV'23 & \XSolidBrush & C & PersFormer & 30 & 57.1 & 100.0 & 28.3 & 98.9 & 63.6\\
\rowcolor{new_blue!3} MapTRv2 & IJCV'24 & \XSolidBrush & C & BEVPool & 24 & 61.4 & 82.5 & 49.9 &  90.3 & 58.8\\
GeMap & ECCV'24 & \XSolidBrush & C & BEVFormer-1 & 24 & 51.3 & 85.4 & 61.2 & 92.3 & 65.6 \\
\rowcolor{new_blue!3} MGMap & CVPR'24 & \XSolidBrush & C & BEVFormer-1 & 24 & 57.9 & 89.7 & 57.5 & 92.4 & 68.1\\
MapQR & ECCV'24 & \XSolidBrush & C & BEVFormer-3 & 24 & 66.4 & 84.9 & 43.8 & 95.4 & 59.0\\

\rowcolor{new_blue!5} MapTR & ICLR'23 & \Checkmark & C & GKT & 24 & 51.3 & 92.8 & 41.0 &  92.8 & 62.3 \\
%\rowcolor{new_orange!5} StreamMapNet & WACV'24 & \Checkmark & C & BEVFormer-1 & 30 & 63.3 & \todo{X} & \todo{X} & \todo{X} & \todo{X}\\
%\rowcolor{new_orange!5} MapTracker & ECCV'24 & \Checkmark & C & BEVFormer-2 & 72 & \todo{X} & \todo{X} & \todo{X} & \todo{X}\\
HRMapNet & ECCV'24 & \Checkmark & C & BEVFormer-1 & 24 & 67.2 & 83.8 & 55.8 & 92.8 & 62.4\\
\bottomrule
\end{tabular}
\end{table}

\subsection{Ablation Study on the Temporal Sampling Interval}
\label{subsec:different_M_experiment}
Temporal sampling serves as the initial step in the stability evaluation benchmark. In the default configuration, the time interval \(M\) is set to 2 to assess the granularity of map changes, as delineated in Table \ref{tab:main_table}. To provide a more comprehensive illustration of our evaluation framework's performance across different temporal sampling intervals, we conducted further experiments with \(M\) values of 3, 5, and 10. The results are presented in Tables \ref{tab:appendix_intreval_1}, \ref{tab:appendix_intreval_2}, and \ref{tab:appendix_intreval_3} respectively. All other experimental settings remain consistent with Table \ref{tab:main_table}.

The ablation studies reveal several important patterns regarding temporal stability assessment. First, as the temporal interval \(M\) increases from 2, 3, 5, 10, most models exhibit a progressive decline in stability scores across all metrics, particularly in Localization Stability. This pattern is consistent across different architectural paradigms and demonstrates the challenge of maintaining consistency over longer time horizons. For instance, MapTR \citep{liao2022maptr} with camera-only input shows a reduction in mAS from 71.6 (\(M\)=2) to 61.9 (\(M\)=10), primarily driven by decreasing Localization Stability. Similar trends are observed for other non-temporal models, with BeMapNet \citep{qiao2023bemapnet} maintaining superior Presence Stability but experiencing significant Localization Stability degradation from 65.8 (\(M\)=2) to 41.6 (\(M\)=10).

The comparative analysis reveals distinctive robustness characteristics across representation paradigms. Models with inherent temporal modeling capabilities, such as StreamMapNet \citep{yuan2024streammapnet} and MapTracker \citep{chen2024maptracker}, demonstrate remarkable resilience to increasing temporal intervals. StreamMapNet maintains exceptional stability with mAS of 93.2 at \(M\)=3, significantly outperforming non-temporal counterparts. This performance advantage is particularly pronounced in Localization Stability, where temporal models consistently exceed 96.0 even at larger intervals, compared to the substantial degradation observed in static models.

The studies also reveal paradigm-specific sensitivity patterns. Geometric-prior-based models like BeMapNet \citep{qiao2023bemapnet} and PivotNet \citep{ding2023pivotnet} maintain perfect Presence Stability across all intervals but exhibit considerable vulnerability in Localization Stability. In contrast, learning-based BEV representation models show more balanced degradation across stability dimensions. The performance variations across intervals provide additional evidence that accuracy (mAP) and stability (mAS) represent independent evaluation dimensions, as models with comparable mAP scores exhibit dramatically different stability characteristics under extended temporal intervals.

These findings underscore the importance of evaluating temporal stability across multiple time scales, as different representation paradigms exhibit distinct degradation patterns. The comprehensive interval analysis reinforces our central thesis that temporal stability constitutes a fundamental performance dimension that requires explicit consideration in online HD mapping system design and evaluation.

% \clearpage

\section{Comprehensive evaluation results with further details}
\label{sec:appendix_6_model_details}
We present a detailed analysis of the performance of various online HD mapping models in terms of both accuracy (mAP) and temporal stability (mAS), based on the comprehensive results summarized in Tables \ref{tab:appendix_table_MapTR} - Table \ref{tab:appendix_table_HRMapNet}. Our analysis highlights how different architectural choices, including backbone networks, temporal modeling, sensor modalities, and BEV encoders, affect these two critical performance dimensions.

\begin{table}[htbp]
\centering
%\fontsize{8pt}{10pt}\selectfont

\caption{Evaluation Results of MapTR}
\fontsize{6pt}{8pt}\selectfont
\label{tab:appendix_table_MapTR}
\begin{tabular}[t]{c|cccc|c|ccc|c|c}
\toprule
\rowcolor{mygray}
\bf Backbone &
\bf Temp &
\bf Modal &
\bf BEV Encoder &
\bf Epoch &
\bf mAP$\uparrow$ &
\bf Presence$\uparrow$ &
\bf Loc$\uparrow$ &
\bf Shape$\uparrow$ &
\bf \bf mAS$\uparrow$ &
\bf Parameters$\downarrow$ 
\\
\midrule
\rowcolor{new_blue!5}
R18 & \XSolidBrush & C & GKT & 24 & 32.4 & 87.8 & 75.0 &  88.5 & 72.8 & \textbf{15.4M}\\

R18 & \XSolidBrush & C & GKT & 110 & 45.5 & 86.0 & 71.7 & \textbf{94.8} & 71.8 & \textbf{15.4M} \\
\rowcolor{new_blue!5}
R50 & \XSolidBrush & C & GKT & 24 & 44.1 & \textbf{91.2} & 65.4 &  90.6 & 71.6 & 36.2M\\

R50 & \XSolidBrush & C & GKT & 110 & 50.5 & 89.8 & 63.2 &  91.0 & 68.2 & 36.2M\\
\rowcolor{new_blue!5}
R50 & \XSolidBrush & C & BEVFormer-1 & 24 & 41.6 & 89.6 & 69.7 &  90.6 & 71.3 & 36.3M\\

R50 & \XSolidBrush & C & BEVPool & 24 & 50.1 & 89.3 & 69.8 &  88.5 & 71.9 & 32.3M\\
\rowcolor{new_blue!5}
R50 & \Checkmark & C & GKT & 24 & 51.3 & 88.6 & 59.7 &  89.3 & 66.6  & 36.2M\\

R50 & \Checkmark & C & BEVFormer-1 & 24 & 53.3 & 90.4 & 69.5 &  91.2 & 73.0  & 36.3M \\
\rowcolor{new_blue!5}
R50 \& Sec. & \XSolidBrush & C \& L & GKT & 24 & \textbf{62.8} & 90.1 & \textbf{75.2} & 90.8 & \textbf{74.9} & 40.1M\\

\bottomrule
\end{tabular}%\rowcolor{new_blue!5}
\end{table}

\subsection{Comprehensive Analysis Towards MapTR}
% As presented in Table \ref{tab:appendix_table_MapTR}, MapTR \citep{liao2022maptr} exhibits a wide range of mAP scores (32.4 to 62.8) and mAS scores (66.6 to 74.9). The model achieves its highest mAP (62.8) and mAS (74.9) when equipped with a ResNet-50 \citep{he2016resnet} backbone, LiDAR-camera fusion, and the GKT \citep{chen2022GKT} BEV encoder. Notably, while temporal integration improves mAP in some configurations (e.g., from 44.1 to 51.3 with the GKT encoder), it can reduce mAS (from 71.6 to 66.6), suggesting that naive temporal fusion may harm stability if not properly aligned with the BEV representation. Additionally, increasing model capacity via a deeper backbone (R50 vs. R18) consistently improves mAP but does not uniformly benefit mAS, indicating a potential trade-off between accuracy and stability.

As presented in Table \ref{tab:appendix_table_MapTR}, the extensive variants of MapTR \citep{liao2022maptr} provide a controlled setting to dissect how distinct representation paradigms influence model behavior. By altering key components while holding the core architecture constant, we can isolate their effects on both accuracy (mAP) and temporal stability (mAS).

\paragraph{Temporal Fusion.} The effect of incorporating temporal fusion is not uniform but is mediated by the underlying BEV representation. When applied to the GKT-based \citep{chen2022GKT} representation, temporal fusion disrupts its core strength. Presence Stability drops from 91.2 to 88.6, and Localization Stability plummets from 65.4 to 59.7, leading to a significant decrease in mAS (71.6 to 66.6). This indicates that the representation formed by GKT is not easily aligned or integrated across time; the temporal module may introduce noise rather than useful context. In contrast, when applied to the BEVFormer-based \citep{li2024bevformer} representation, which is already designed for spatiotemporal modeling, temporal fusion acts as a complementary enhancement. It improves mAP substantially (41.6 to 53.3) while slightly improving or maintaining stability scores, resulting in a higher mAS (71.3 to 73.0). This demonstrates that temporal fusion is most effective when the base representation is inherently compatible with processing sequential data. The effectiveness of temporal fusion is not a standalone property but is contingent on the representational capacity of the BEV encoder. It amplifies the capabilities of a temporally-aware representation (BEVFormer) but can degrade the performance of a primarily spatially-focused one (GKT). Therefore, temporal fusion serves as a force multiplier for representations already predisposed to temporal modeling, but can be detrimental to those that are not.

\paragraph{2D Backbone.} Changing the 2D backbone from ResNet-18 to ResNet-50 shifts the model's representational focus towards more complex visual patterns. This shift has a clear effect: it consistently improves mAP (32.4 to 44.1) by leveraging higher-capacity feature extraction. However, this comes with a redistribution of stability properties: Presence Stability improves (87.8 to 91.2), but Localization Stability worsens significantly (75.0 to 65.4). The deeper network appears to learn a representation that is more sensitive to semantic content but potentially more susceptible to per-frame variations in texture or lighting, which can harm geometric consistency. The backbone network influences the type of features that form the basis of the map representation. More powerful backbones enhance semantic discrimination but can introduce high-frequency noise that undermines geometric stability, suggesting that representations favoring stability may require features that are invariant to superficial appearance changes.

\begin{table}[htbp]
\centering
%\fontsize{8pt}{10pt}\selectfont

\caption{Evaluation Results of BeMapNet}
\fontsize{6pt}{8pt}\selectfont
\label{tab:appendix_table_BeMapNet}
\begin{tabular}[t]{c|cccc|c|ccc|c|c}
\toprule
\rowcolor{mygray}
\bf Backbone &
\bf Temp &
\bf Modal &
\bf BEV Encoder &
\bf Epoch &
\bf mAP$\uparrow$ &
\bf Presence$\uparrow$ &
\bf Loc$\uparrow$ &
\bf Shape$\uparrow$ &
\bf \bf mAS$\uparrow$ &
\bf Parameters$\downarrow$ 
\\
\midrule
\rowcolor{new_blue!5}
Effb0 & \XSolidBrush & C & IPM-PE & 30 & 60.7 & \textbf{100.0} & \textbf{67.9} &  97.9 & \textbf{82.9} & 55.4M\\

R50 & \XSolidBrush & C & IPM-PE & 30 & 61.4 & \textbf{100.0} & 65.8 &  97.9 & 81.9 & 73.8M\\
\rowcolor{new_blue!5}
SwinT & \XSolidBrush & C & IPM-PE & 30 & 64.1 & \textbf{100.0} & 62.8 &  98.0 & 80.4 & 79.6M\\

R50 & \XSolidBrush & C & IPM-PE & 110 & 66.2 & \textbf{100.0} & 62.1 & \textbf{98.2} & 80.2 & 73.8M\\
\rowcolor{new_blue!5}
SwinT & \XSolidBrush & C & IPM-PE & 110 & \textbf{68.3} & \textbf{100.0} & 64.0 & \textbf{98.2} & 81.1 & 79.6M\\

\bottomrule
\end{tabular}%\rowcolor{new_blue!5}
\end{table}
\begin{table}[htbp]
\centering
%\fontsize{8pt}{10pt}\selectfont

\caption{Evaluation Results of PivotNet}
\fontsize{6pt}{8pt}\selectfont
\label{tab:appendix_table_PivotNet}
\begin{tabular}[t]{c|cccc|c|ccc|c|c}
\toprule
\rowcolor{mygray}
\bf Backbone &
\bf Temp &
\bf Modal &
\bf BEV Encoder &
\bf Epoch &
\bf mAP$\uparrow$ &
\bf Presence$\uparrow$ &
\bf Loc$\uparrow$ &
\bf Shape$\uparrow$ &
\bf \bf mAS$\uparrow$ &
\bf Parameters$\downarrow$ 
\\
\midrule
\rowcolor{new_blue!5}
Effb0 & \XSolidBrush & C & PersFormer & 30 & 57.8 & \textbf{100.0} & 71.8 & 97.2 & 84.5 & \textbf{17.1M}\\

R50 & \XSolidBrush & C & PersFormer & 30 & 57.1 & \textbf{100.0} & 71.4 & 97.2 & 84.3 & 41.2M\\
\rowcolor{new_blue!5}
Swin-T & \XSolidBrush & C & PersFormer & 30 & 61.6 & \textbf{100.0} & 71.6 & 97.2 & 84.4 & 44.8M\\

Swin-T & \XSolidBrush & C & PersFormer & 110 & \textbf{66.4} & \textbf{100.0} & \textbf{72.1} & \textbf{97.4} & \textbf{84.8} & 44.8M\\

\bottomrule
\end{tabular}%\rowcolor{new_blue!5}
\end{table}

\subsection{In-Depth Analysis of BeMapNet and PivotNet}

As shown in Table \ref{tab:appendix_table_BeMapNet} and Table \ref{tab:appendix_table_PivotNet}, both BeMapNet\citep{qiao2023bemapnet} and PivotNet\citep{ding2023pivotnet} demonstrate stable mAP and mAS performance, with BeMapNet achieving mAP scores of 60.7 to 68.3 and mAS values of 80.2 to 82.9, while PivotNet demonstrates mAP values of 57.1 to 66.4 and mAS scores of 84.3 to 84.8.This indicats that these two models are insensitive to backbone network selection and training epoch configurations. The difference lies in the fact that PivotNet achieves its highest mAS when using Swin-T\citep{liu2021swintransformer} as the backbone, while BeMapNet attains its peak mAS value with an EfficientNet-B0\citep{tan2019efficientnet} backbone. 

It should be specifically noted that both BeMapNet and PivotNet adopt a “dynamic vectorized sequence" representation for map encoding, which explains their consistently perfect presence metrics (100\%). However, this representation format severely limits the localization stability of map instances, resulting in significantly lower performance compared to models like GeMap\citep{zhang2024gemap}, MGMap\citep{liu2024mgmap}, and MapQR\citep{liu2024mapqr}.

\begin{table}[htbp]
\centering
%\fontsize{8pt}{10pt}\selectfont

\caption{Evaluation Results of MapTRv2}
\fontsize{6pt}{8pt}\selectfont
\label{tab:appendix_table_MapTRv2}
\begin{tabular}[t]{c|cccc|c|ccc|c|c}
\toprule
\rowcolor{mygray}
\bf Backbone &
\bf Temp &
\bf Modal &
\bf BEV Encoder &
\bf Epoch &
\bf mAP$\uparrow$ &
\bf Presence$\uparrow$ &
\bf Loc$\uparrow$ &
\bf Shape$\uparrow$ &
\bf \bf mAS$\uparrow$ &
\bf Parameters$\downarrow$ 
\\
\midrule
\rowcolor{new_blue!5}
R18 & \XSolidBrush & C & BEVPool & 24 & 57.2 & 91.0 & \textbf{73.2} &  \textbf{91.2} & \textbf{75.6} & \textbf{27.9M}\\

R50 & \XSolidBrush & C & BEVPool & 24 & \textbf{61.4} & \textbf{91.5} & 68.6 &  91.0 & 74.0 & 40.6M\\

\bottomrule
\end{tabular}%\rowcolor{new_blue!5}
\end{table}

\subsection{Evaluation and Analysis of MapTRv2}
MapTRv2 \citep{liao2025maptrv2} improves upon MapTR \citep{liao2022maptr} with higher baseline mAP (57.2–61.4) and mAS (74.0–75.6). Interestingly, the R18 backbone achieves higher mAS (75.6) than R50 (74.0), despite a lower mAP (57.2 \textit{vs.} 61.4), as illustrated in Table \ref{tab:appendix_table_MapTR} and Table \ref{tab:appendix_table_MapTRv2}, reinforcing the independence of accuracy and stability.

\begin{table}[htbp]
\centering
%\fontsize{8pt}{10pt}\selectfont

\caption{Evaluation Results of GeMap}
\fontsize{6pt}{8pt}\selectfont
\label{tab:appendix_table_GeMap}
\begin{tabular}[t]{c|cccc|c|ccc|c|c}
\toprule
\rowcolor{mygray}
\bf Backbone &
\bf Temp &
\bf Modal &
\bf BEV Encoder &
\bf Epoch &
\bf mAP$\uparrow$ &
\bf Presence$\uparrow$ &
\bf Loc$\uparrow$ &
\bf Shape$\uparrow$ &
\bf \bf mAS$\uparrow$ &
\bf Parameters$\downarrow$ 
\\
\midrule
\rowcolor{new_blue!5}
R50 & \XSolidBrush & C & BEVFormer-1 & 24 & 51.3 & 92.3 & 69.7 & 92.6 & 75.5 & \textbf{44.1M}\\

R50 & \XSolidBrush & C & BEVFormer-1 & 110 & 62.7 & 91.1 & 67.5 & \textbf{94.5} & 74.7 & \textbf{44.1M}\\
\rowcolor{new_blue!5}
Swin-T & \XSolidBrush & C & BEVFormer-1 & 110 & 72.0 & 92.2 & \textbf{74.9} & 93.2 & \textbf{78.1} & 50.5M\\

V2-99 & \XSolidBrush & C & BEVFormer-1 & 110 & 72.0 & 89.2 & 71.5 & 92.6 & 74.2 & 92.6M\\
\rowcolor{new_blue!5}
V2-99(DD3D) & \XSolidBrush & C & BEVFormer-1 & 110 & \textbf{76.0} & \textbf{93.4} & 66.9 &  93.7 & 75.1 & 92.6M\\

R50 \& second & \XSolidBrush & C \& L & BEVFormer-1& 110 & 66.5 & 89.1 & 66.3 &  92.7 & 71.8 & 48.0M\\

\bottomrule
\end{tabular}%\rowcolor{new_blue!5}
\end{table}

\subsection{Discussion on GeMap}

As illustrated in Table \ref{tab:appendix_table_GeMap}, GeMap \citep{zhang2024gemap} presents a particularly instructive case for examining the complex relationship between accuracy and stability. The model demonstrates a strong capacity for high accuracy, with its mAP score scaling significantly from 51.3 to a top score of 76.0 when employing a powerful V2-99 \citep{lee2019v299} backbone and extended training. However, this pursuit of accuracy often introduces instability, as evidenced by its mAS scores, which range from a moderate 71.8 to a more competitive 78.1.

A critical observation is the divergent effect of LiDAR fusion. While integrating LiDAR data with a ResNet-50 \citep{he2016resnet} backbone yields a predictable improvement in mAP (+6.1\%, from 62.7 to 66.5), it conversely leads to a decrease in mAS (-3.9\%, from 74.7 to 71.8). This result challenges the conventional wisdom that more sensor data invariably leads to more robust perception. It suggests that GeMap's architecture, while effectively leveraging LiDAR for geometric precision in a single frame, may lack the necessary mechanisms to harmonize the potentially noisy or asynchronous multi-modal signals across time, leading to increased jitter or flickering.

Furthermore, the model exhibits high sensitivity to backbone design. The Swin-T \citep{liu2021swintransformer} backbone strikes the most favorable balance, achieving the highest mAS (78.1) alongside a high mAP (72.0). In contrast, the larger V2-99 \citep{lee2019v299} backbone, despite achieving the peak mAP (76.0), produces a lower mAS (75.1). The degradation in Localization Stability (from 74.9 with Swin-T \citep{liu2021swintransformer} to 66.9 with V2-99 \citep{lee2019v299}) is especially notable, implying that the increased representational power of the larger backbone may overfit to single frame features at the expense of temporal coherence. This pattern underscores that for stability, simply scaling up model capacity is not a sufficient strategy and may even be counterproductive without explicit temporal regularization.

\begin{table}[htbp]
\centering
%\fontsize{8pt}{10pt}\selectfont

\caption{Evaluation Results of MGMap}
\fontsize{6pt}{8pt}\selectfont
\label{tab:appendix_table_MGMap}
\begin{tabular}[t]{c|cccc|c|ccc|c|c}
\toprule
\rowcolor{mygray}
\bf Backbone &
\bf Temp &
\bf Modal &
\bf BEV Encoder &
\bf Epoch &
\bf mAP$\uparrow$ &
\bf Presence$\uparrow$ &
\bf Loc$\uparrow$ &
\bf Shape$\uparrow$ &
\bf \bf mAS$\uparrow$ &
\bf Parameters$\downarrow$ 
\\
\midrule
\rowcolor{new_blue!5}
R50 & \XSolidBrush & C & BEVFormer-1 & 24 & 58.0 & 92.2 & 75.0 &  92.3 & 78.0 & 55.9M\\

\bottomrule
\end{tabular}%\rowcolor{new_blue!5}
\end{table}

\subsection{Analysis for MGMap}
MGMap \citep{liu2024mgmap} achieves a balanced profile (mAP: 58.0, mAS: 78.0) with a ResNet-50 backbone \citep{he2016resnet} and BEVFormer \citep{li2024bevformer} encoder, as presented in Table \ref{tab:appendix_table_MGMap}. Its strong Localization and Shape Stability scores (75.0 and 92.3, respectively) suggest robustness against geometric jitter.

\begin{table}[htbp]
\centering
%\fontsize{8pt}{10pt}\selectfont

\caption{Evaluation Results of MapQR}
\fontsize{6pt}{8pt}\selectfont
\label{tab:appendix_table_MapQR}
\begin{tabular}[t]{c|cccc|c|ccc|c|c}
\toprule
\rowcolor{mygray}
\bf Backbone &
\bf Temp &
\bf Modal &
\bf BEV Encoder &
\bf Epoch &
\bf mAP$\uparrow$ &
\bf Presence$\uparrow$ &
\bf Loc$\uparrow$ &
\bf Shape$\uparrow$ &
\bf \bf mAS$\uparrow$ &
\bf Parameters$\downarrow$ 
\\
\midrule
\rowcolor{new_blue!5}
R18 & \XSolidBrush & C & BEVFormer-3 & 24 & 62.3 & 88.2 & 73.1 &  92.5 & 74.1 & \textbf{112.6M} \\

R50 & \XSolidBrush & C & BEVFormer-3 & 24 & 66.4 & 91.8 & 75.6 &  91.6 & 77.8 & 125.4M\\
\rowcolor{new_blue!5}
R50 & \XSolidBrush & C & BEVFormer-3 & 110 & \textbf{72.6} & \textbf{92.4} & \textbf{75.9} & \textbf{96.4} & \textbf{80.3} & 125.4M\\

\bottomrule
\end{tabular}%\rowcolor{new_blue!5}
\end{table}

\subsection{Study of MapQR}
As shown in Table \ref{tab:appendix_table_MapQR}, MapQR \citep{liu2024mapqr} shows a clear positive scaling trend: larger backbones and longer training improve both mAP (62.3 to 72.6) and mAS (74.1 to 80.3). This indicates that the model’s architecture supports stable learning under increased capacity.

\begin{table}[htbp]
\centering
%\fontsize{8pt}{10pt}\selectfont

\caption{Evaluation Results of StreamMapNet}
\fontsize{6pt}{8pt}\selectfont
\label{tab:appendix_table_StreamMapNet}
\begin{tabular}[t]{c|cccc|c|ccc|c|c}
\toprule
\rowcolor{mygray}
\bf Backbone &
\bf Temp &
\bf Modal &
\bf BEV Encoder &
\bf Epoch &
\bf mAP$\uparrow$ &
\bf Presence$\uparrow$ &
\bf Loc$\uparrow$ &
\bf Shape$\uparrow$ &
\bf \bf mAS$\uparrow$ &
\bf Parameters$\downarrow$ 
\\
\midrule
\rowcolor{new_blue!5}
R50 & \XSolidBrush & C & BEVFormer-1 & 30 & 51.7 & 87.0 & 97.8 &  95.1 & 83.8 & 56.0M\\

R18 & \Checkmark & C & BEVFormer-1 & 30 & 27.8 & 87.1 & 98.4 &  94.6 & 85.0 & \textbf{42.5M}\\
\rowcolor{new_blue!5}
R50 & \Checkmark & C & BEVFormer-1 & 30 & \textbf{63.4} & 96.6 & 97.7 & 92.3 & 91.9 & 56.3M\\

R50 & \Checkmark & C & BEVFormer-1 & 24 & 51.2 & \textbf{97.0} & \textbf{98.5} & \textbf{96.1} & \textbf{94.4} & 56.3M\\

\bottomrule
\end{tabular}%\rowcolor{new_blue!5}
\end{table}
\subsection{Towards a Comprehensive Analysis of StreamMapNet}
StreamMapNet \citep{yuan2024streammapnet} stands out as the paradigm for temporally stable online mapping, achieving the highest mAS scores in our benchmark, ranging from 83.8 to an exceptional 94.4. This performance is primarily driven by its native temporal architecture, which is explicitly designed to model consistency across frames.

The most striking feature of StreamMapNet is its near-perfect Localization Stability (97.7–98.5), the highest among all models, as shown in Table \ref{tab:main_table} and Table \ref{tab:appendix_table_StreamMapNet}. This indicates an exceptional ability to suppress the positional jitter of map elements over time, a critical factor for downstream planning tasks. The analysis clearly shows that temporal fusion is not merely an optional add-on but the core determinant of its performance. Enabling temporal modeling (comparing the R50, \Checkmark vs. R50, \XSolidBrush configurations) results in a dramatic improvement in both mAP (+22.6\%, from 51.7 to 63.4) and mAS (+9.7\%, from 83.8 to 91.9). This dual improvement confirms that effectively leveraging historical context can simultaneously enhance per-frame accuracy and inter-frame consistency.

\begin{table}[htbp]
\centering
%\fontsize{8pt}{10pt}\selectfont

\caption{Evaluation Results of MapTracker}
\fontsize{6pt}{8pt}\selectfont
\label{tab:appendix_table_MapTracker}
\begin{tabular}[t]{c|cccc|c|ccc|c|c}
\toprule
\rowcolor{mygray}
\bf Backbone &
\bf Temp &
\bf Modal &
\bf BEV Encoder &
\bf Epoch &
\bf mAP$\uparrow$ &
\bf Presence$\uparrow$ &
\bf Loc$\uparrow$ &
\bf Shape$\uparrow$ &
\bf \bf mAS$\uparrow$ &
\bf Parameters$\downarrow$ 
\\
\midrule
\rowcolor{new_blue!5}
R18 & \XSolidBrush & C & BEVFormer-2 & 72 & 62.8 & \textbf{95.3} & 97.3 & 85.9 & 87.4 & \textbf{60.7M}\\

R50 & \XSolidBrush & C & BEVFormer-2 & 72 & 68.3 & 94.5 & 97.9 & 93.8 & 90.8 & 74.0M\\
\rowcolor{new_blue!5}
R18 & \Checkmark & C & BEVFormer-2 & 48 & 69.3 & 94.8 & 98.2 & 94.8 & 91.5 & \textbf{60.7M}\\

R18 & \Checkmark & C & BEVFormer-2 & 72 & 71.9 & 92.9 & \textbf{98.5} & 94.8 & 89.9 & \textbf{60.7M} \\
\rowcolor{new_blue!5}
R50 & \Checkmark & C & BEVFormer-2 & 48 & 73.0 & 91.7 & \textbf{98.5} & \textbf{96.0} & \textbf{91.7} & 74.0M \\

R50 & \Checkmark & C & BEVFormer-2 & 72 & \textbf{76.0} & 93.3 & 98.1 & 95.8 & 90.4 & 74.0M \\

\bottomrule
\end{tabular}%\rowcolor{new_blue!5}
\end{table}

\subsection{A Comprehensive Analysis of MapTracker}
MapTracker \citep{chen2024maptracker} represents another strong temporal model that successfully balances state-of-the-art accuracy with high stability, which is shown in Table \ref{tab:main_table} and Table \ref{tab:appendix_table_MapTracker}. It achieves the highest overall mAP (76.0) in our benchmark while maintaining mAS scores above 87.4, peaking at 91.7.

Similar to StreamMapNet \citep{yuan2024streammapnet}, MapTracker's integration of temporal fusion (“Temp = \Checkmark") consistently boosts mAP (e.g., from 62.8 to 69.3 for R18) while preserving high mAS. This reinforces the conclusion that architectures designed with temporal reasoning in mind from the ground up are essential for high-performance online mapping. The model also exhibits very strong Localization and Shape Stability, often exceeding 98.0 and 94.0, respectively, which is characteristic of models that effectively aggregate information over time.

However, MapTracker reveals a nuanced trade-off related to training duration. For both the R18 and R50 backbones, extending training from 48 to 72 epochs leads to a further increase in mAP but a slight decrease in mAS (e.g., R50: mAP 73.0 to 76.0, mAS 91.7 to 90.4). This pattern, which we term optimization sensitivity, suggests that as the model continues to minimize a primarily accuracy-oriented loss function, it may gradually overfit to single-frame details, thereby sacrificing some temporal smoothness. This highlights a key area for future research: the development of loss functions or regularization techniques that explicitly penalize temporal instability during training to prevent this erosion.

\begin{table}[htbp]
\centering
%\fontsize{8pt}{10pt}\selectfont

\caption{Evaluation Results of HRMapNet}
\fontsize{5.5pt}{8pt}\selectfont
\label{tab:appendix_table_HRMapNet}
\begin{tabular}[t]{c|cc|ccc|c|ccc|c|c}
\toprule
\rowcolor{mygray}
\bf Backbone &
\bf Temp &
\bf Initial Map &
\bf Modal &
\bf BEV Encoder &
\bf Epoch &
\bf mAP$\uparrow$ &
\bf Presence$\uparrow$ &
\bf Loc$\uparrow$ &
\bf Shape$\uparrow$ &
\bf \bf mAS$\uparrow$ &
\bf Parameters$\downarrow$ 
\\
\midrule
\rowcolor{new_blue!5}
R50 & \Checkmark & \XSolidBrush & C & BEVFormer-1 & 24 & 67.2 & 92.3 & 70.5 & 91.5 & 75.9 & \textbf{47.3M}\\

R50 & \Checkmark & Testing Map & C & BEVFormer-1 & 24 & 73.0 & \textbf{94.9} & 71.4 & 93.0 & \textbf{78.4} & \textbf{47.3M}\\
\rowcolor{new_blue!5}
R50 & \Checkmark & Training Map & C & BEVFormer-1 & 24 & \textbf{83.6} & 89.9 & \textbf{75.9} & \textbf{93.2} & 76.7 & \textbf{47.3M}\\

R50 & \Checkmark & \XSolidBrush & C & BEVFormer-1 & 110 & 73.5 & 90.5 & 74.1 & 92.7 & 75.9 & \textbf{47.3M} \\

\bottomrule
\end{tabular}%\rowcolor{new_blue!5}
\end{table}

\subsection{Detailed Investigation of HRMapNet}
As presented in Table \ref{tab:appendix_table_HRMapNet}, HRMapNet \citep{zhang2024hrmapnet} incorporates a distinct representation paradigm by integrating static map priors into a temporal mapping framework. The performance variations across its configurations provide critical insights into the interaction between dynamic sensory input and static prior knowledge.

The most pronounced effect is observed on single frame accuracy. Utilizing a map prior during training yields a substantial improvement in mAP, elevating the score from 67.2 to 83.6. This result indicates that the model's representation effectively internalizes the structural constraints provided by the high quality offline map, leading to superior geometric precision in individual frames.

In contrast, the impact of map priors on temporal stability is more complex and less direct. The configuration employing a prior only during testing achieves the highest mAS of 78.4 and the highest Presence Stability of 94.9. This suggests that an externally provided prior can serve as a stabilizing reference during inference, enhancing detection consistency without being fully baked into the model parameters.

However, when the model is trained with the map prior, a different pattern emerges. While this configuration achieves the highest mAP, its mAS of 76.7 is lower than the testing prior variant. Notably, its Presence Stability decreases to 89.9. This indicates that deep integration of the static prior during training may lead to a representation that is overly reliant on persistent features, potentially at the expense of robustness to real world variations encountered in a temporal sequence. The model may become less adept at handling cases where the prior is imperfect or where dynamic scenes deviate from the stored map.

Furthermore, extending training to 110 epochs without any initial map prior improves mAP to 73.5 but leaves mAS unchanged at 75.9. This stability saturation effect underscores that prolonged training on a single frame accuracy objective has diminishing returns for temporal consistency. The gain in stability achieved through the intelligent use of a testing time prior surpasses that achieved by simply training the baseline model longer.

In summary, HRMapNet demonstrates that static map priors constitute a powerful representation for enhancing perceptual accuracy. Their utility for improving temporal stability, however, is contingent on the method of integration. A prior used as a dynamic guidance signal at inference can bolster consistency, whereas deeply embedding the prior into the model through training may introduce a trade off, favoring accuracy over stability. This highlights that the effective fusion of dynamic and static representations remains a key challenge for robust online mapping.

\subsection{Summary of Comprehensive Evaluation}
The comprehensive evaluation of ten representative online HD mapping models demonstrates that different representation paradigms induce distinct performance characteristics along the accuracy-stability spectrum. Our analysis reveals that these two performance dimensions are independently influenced by specific architectural choices and their underlying representational biases.

Models incorporating strong geometric priors, such as BeMapNet \citep{qiao2023bemapnet} and PivotNet \citep{ding2023pivotnet}, achieve exceptional Presence Stability due to their structure-aware representations. In contrast, architectures based on learned view transformations like BEVFormer \citep{li2024bevformer} exhibit superior Localization Stability, benefiting from their spatially coherent bird's eye view representations.

Temporal modeling effectiveness shows fundamental dependence on representational compatibility. Architectures with native temporal designs demonstrate that explicit sequence modeling produces the highest stability scores. However, the integration of temporal modules requires careful alignment with the base representation, as evidenced by the varied outcomes when adding temporal components to different BEV encoders.

Multi-modal integration exhibits model-dependent effects on stability. While sensor fusion generally enhances accuracy, its impact on temporal consistency varies across architectures, indicating that effective multi-modal representation requires specialized design beyond simple feature combination.

The relationship between model capacity and performance reveals consistent patterns. Larger backbones produce substantial accuracy gains but yield inconsistent effects on stability, suggesting that representational capacity alone cannot address temporal consistency requirements.

The comparison between static priors and dynamic modeling highlights their complementary roles. While static priors significantly boost accuracy, dynamic temporal modeling proves essential for achieving temporal stability, indicating that these two approaches address distinct aspects of the mapping problem.

These findings collectively suggest that accuracy and stability are governed by different aspects of representation design. This understanding points to the need for future architectures that can simultaneously support high-fidelity spatial representation and robust temporal consistency through integrated design principles.

% \clearpage

\section{Supplemental Visualization and Analysis}
\label{sec:appendix_7_vis}

\begin{figure}[htbp]
    \centering
    \includegraphics[width=0.9\linewidth]{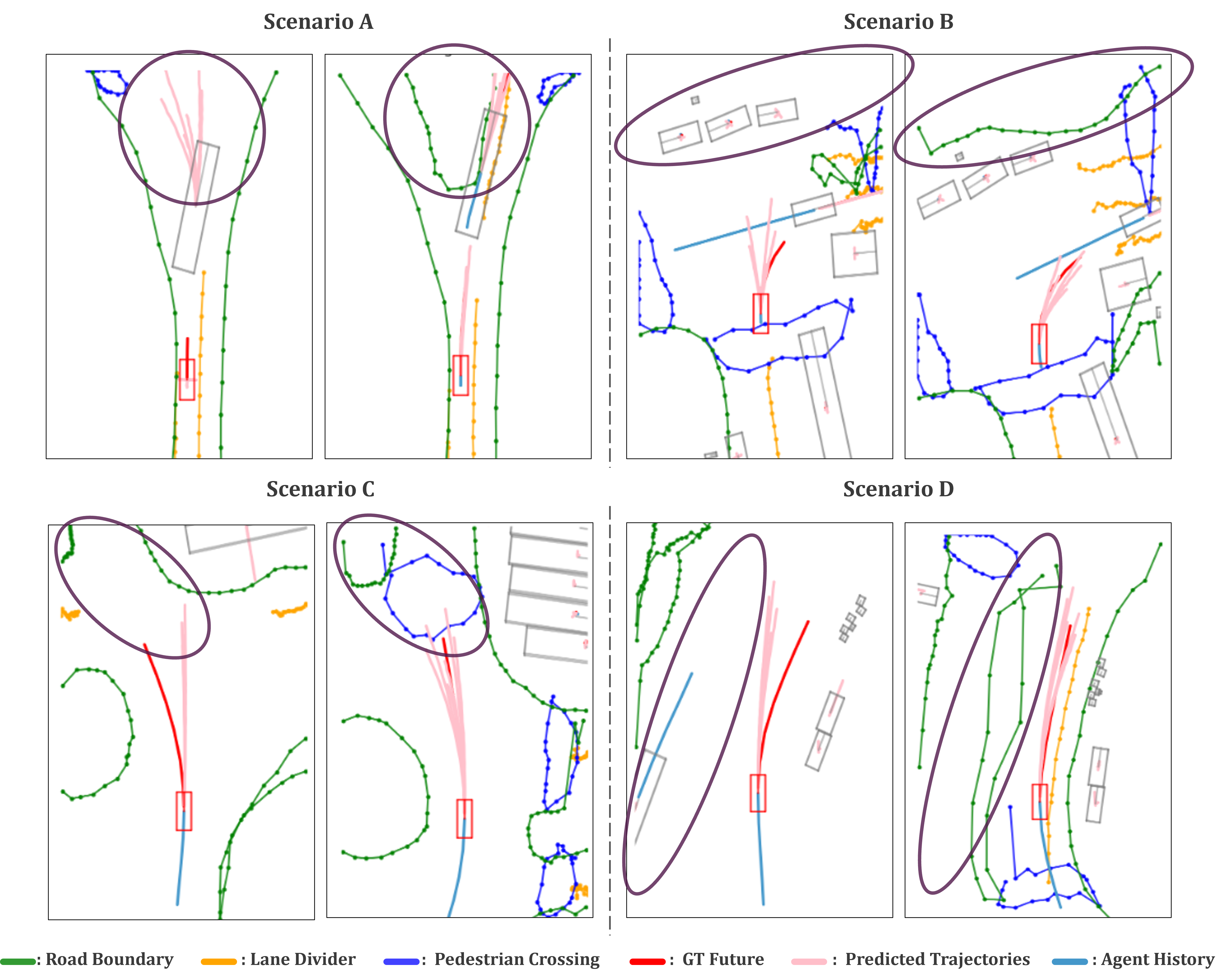}
    \caption{\textbf{Impact of Temporal Inconsistency in Map Element Presence on Downstream Tasks}}
    \label{fig:appendix_downstream_1}
\end{figure}

\begin{figure}[htbp]
    \centering
    \includegraphics[width=0.6\linewidth]{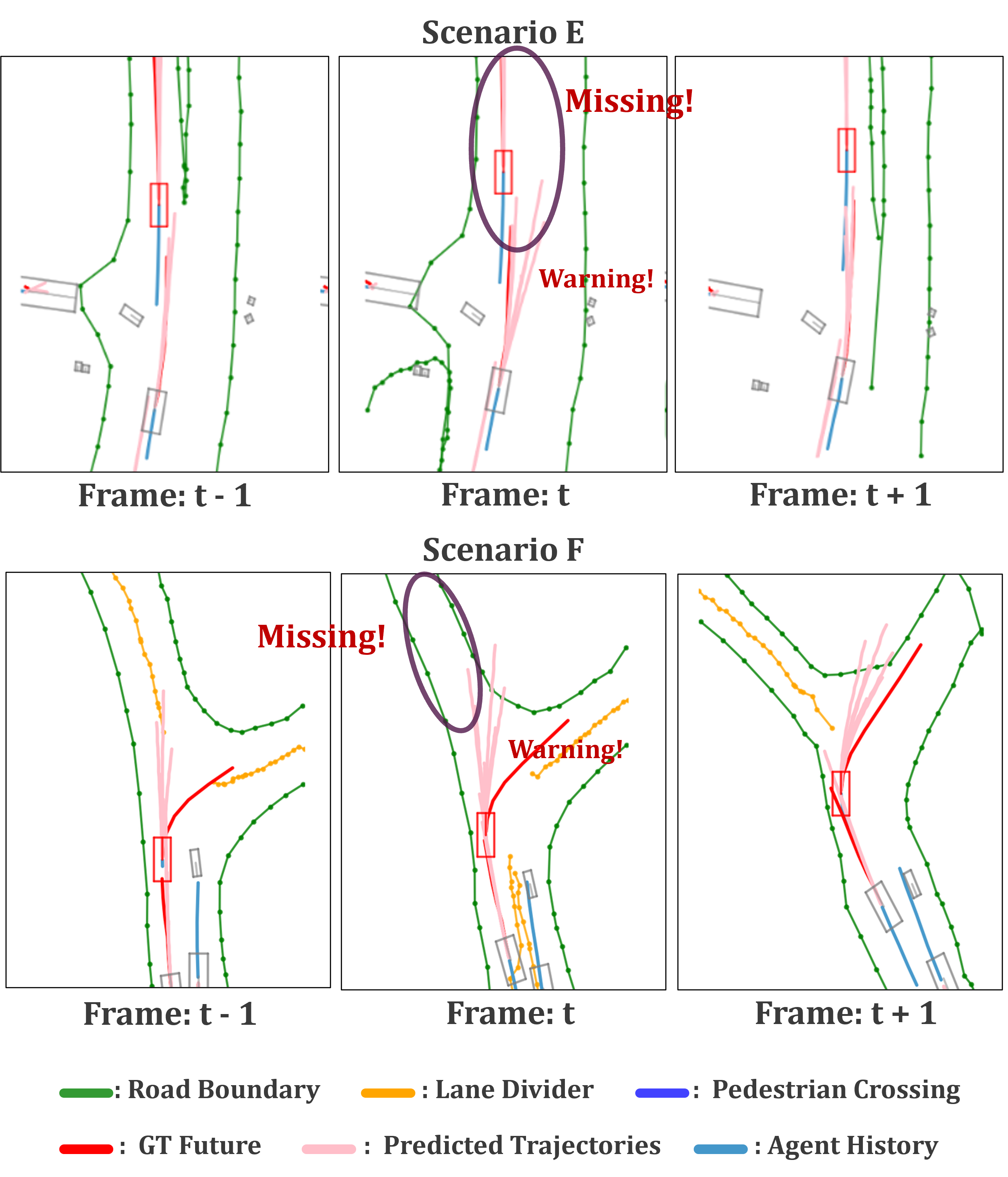}
    \caption{\textbf{Impact of Flickering in Predicted Map Elements on Downstream Tasks.}}
    \label{fig:appendix_downstream_2}
\end{figure}

\subsection{Impact of Unstable Map Predictions on Downstream Tasks}
In this part, we visualize how temporal fluctuations in online mapping predictions impact downstream tasks, as shown in Figures \ref{fig:appendix_downstream_1} and \ref{fig:appendix_downstream_2}. 

Figure \ref{fig:appendix_downstream_1} illustrates the effect of map changes between consecutive frames on these tasks. 

In Scenario A at time t, the ego vehicle fails to detect an intersection ahead due to occlusion by a leading vehicle, leading it to predict the vehicle will turn left. At time t+1, the ego vehicle successfully identifies the intersection, resulting in a corrected prediction of the leading vehicle's trajectory.

In Scenario B, occlusion by three vehicles directly ahead prevents the ego vehicle from detecting the road boundary behind them, causing it to plan a straight path that would collide with the curb. At time t+1, after moving forward, the ego vehicle observes the previously hidden map element and correctly plans a right turn.

In Scenario C at time t, the vehicle does not observe the crosswalk ahead and plans to continue straight. After advancing and detecting the map element, the ego vehicle adjusts its plan accordingly.

In Scenario D's initial frame, occlusion by other vehicles prevents the ego vehicle from predicting the road boundary to its left, leading to a straight path plan that risks a curb collision. Upon detecting the map element on the left in the next frame, it adjusts its trajectory for safe navigation.

Based on the analysis of Figure \ref{fig:appendix_downstream_1}, a key conclusion can be drawn: map elements are critical for autonomous systems to perform downstream tasks such as trajectory prediction and planning. Temporal instability in the perception of these elements can lead to significantly different and potentially unsafe predictions and plans.

As shown in Figure \ref{fig:appendix_downstream_2}, we provide further visualization of how flickering map elements impact downstream tasks over time.

In Scenario E at frame t-1, the ego vehicle is proceeding normally. However, at frame t, a flicker occurs in the predicted road boundary to the right of the lead vehicle, caused by instability in the online mapping model. This leads the ego vehicle to perceive an opportunity to overtake on the right, resulting in a planning decision to steer right and attempt a pass. By frame t+1, the model correctly perceives the road boundary again, causing the ego vehicle to abort the maneuver and resume a straight path.

In Scenario F at frame t-1, the ego vehicle observes the lane divider ahead and plans a normal trajectory. At frame t, however, the predicted lane divider suddenly disappears, causing the planning module to become uncertain and unable to confidently decide between a lane change or continuing straight.

\subsection{mAP \textit{vs.} mAS}

\begin{figure}[htbp]
    \centering
    \includegraphics[width=0.7\linewidth]{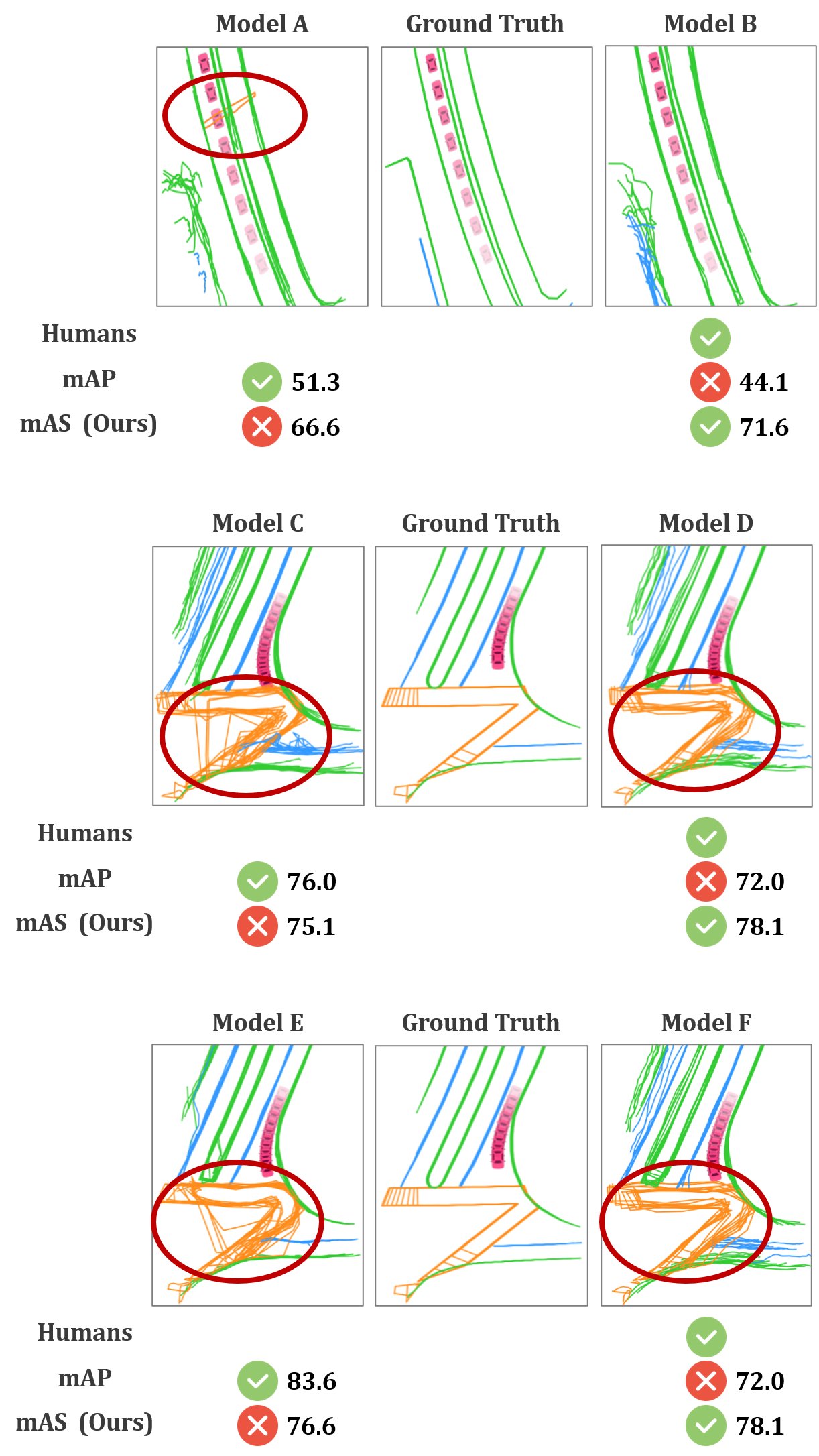}
    \caption{\textbf{Evaluating trustworthiness of online mapping models using human judgment, traditional mAP, and our mAS metric.}}
    \label{fig:mAP_vs_mAS_fig1}
\end{figure}

\begin{figure}[htbp]
    \centering
    \includegraphics[width=0.7\linewidth]{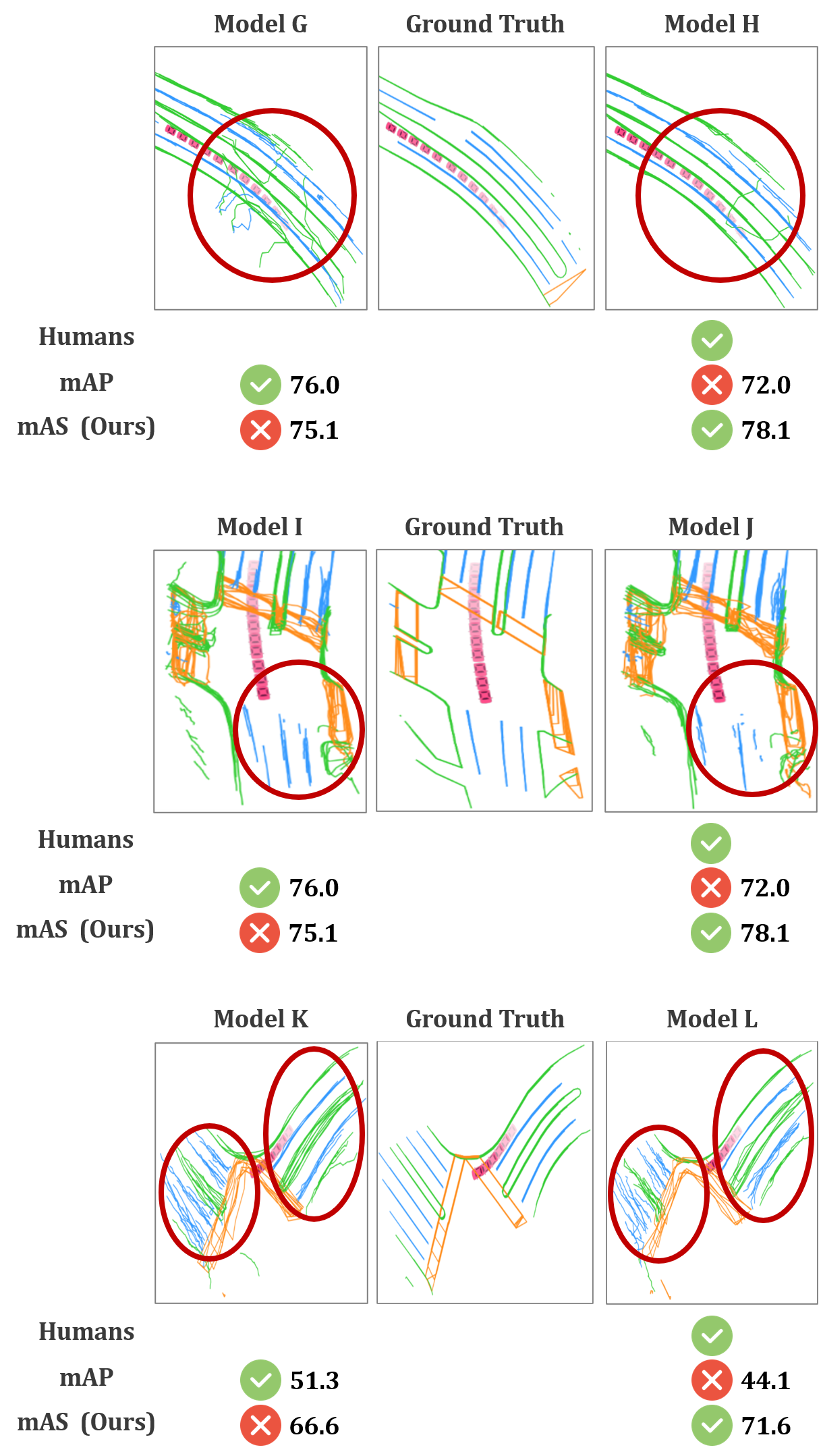}
    \caption{\textbf{Evaluating trustworthiness of online mapping models using human judgment, traditional mAP, and our mAS metric.}}
    \label{fig:mAP_vs_mAS_fig2}
\end{figure}

In this section, we present additional cases where mAP proves to be a misleading indicator for evaluating temporal stability, whereas our proposed mAS correctly assesses temporal stability, as demonstrated in Figure \ref{fig:mAP_vs_mAS_fig1} and Figure \ref{fig:mAP_vs_mAS_fig2}. These examples clearly show that mAP should not be used as a criterion for temporal stability evaluation, whereas mAS provides a more accurate assessment of temporal stability.

% In Figure \ref{fig:mAP_vs_mAS_fig1} and Figure \ref{fig:mAP_vs_mAS_fig2}, Model A and Model K represent the MapTR model integrated with temporal features, employing a ResNet-50 backbone and GKT encoder, trained for 24 epochs. Model B and Model L represent the MapTR model without temporal features, utilizing the same ResNet-50 backbone and GKT encoder, also trained for 24 epochs. Model C,Model G and Model I  represent the GeMap model, employing a vov99-dd3d backbone and BevFormer encoder, trained for 24 epochs.
% Model A represents the MapTR model integrated with temporal features, employing a ResNet-50 backbone and GKT encoder, trained for 24 epochs.Model B represents the MapTR model without temporal features, utilizing the same ResNet-50 backbone and GKT encoder, also trained for 24 epochs. While Model A demonstrates superior metric performance compared to Model B, our mAS evaluation reveals that Model A exhibits inferior stability. Visual comparison of their outputs shows that Model A occasionally detects non-existent crosswalks in individual frames—a manifestation of poor field stability. This observation confirms that Model A's stability is indeed weaker than Model B's, consistent with the mAS evaluation results.

%----A，B-------
As shown in Figure \ref{fig:mAP_vs_mAS_fig1}, model A represents the MapTR \citep{liao2022maptr} model integrated with temporal features, employing a ResNet-50 backbone \citep{he2016resnet} and GKT encoder \citep{chen2022GKT}, trained for 24 epochs. Model B represent the MapTR \cite{liao2022maptr} model without temporal features, utilizing the same ResNet-50 backbone \citep{he2016resnet} and GKT encoder \citep{chen2022GKT}, also trained for 24 epochs. Although model A achieves a relatively high mAP of 51.3, its stability is inferior to that of model B. Specifically, in the visualization results of model A, a crosswalk flickers into view on the road, and additionally, the leftmost lane divider visualized by model A flickers frequently. In contrast, although model B has a lower mAP compared to model A, it does not produce sudden flickering of other map elements in the middle of the road and is able to consistently predict the lane divider on the roadside in nearly every frame.
As illustrated in Figure \ref{fig:mAP_vs_mAS_fig1}, both model C and model D represent GeMap \citep{zhang2024gemap} models. Model C was trained for 110 epochs using the Swin-T \citep{liu2021swintransformer} backbone network and the BEVFormer encoder. Model D was trained for 110 epochs using the V2-99 (DD3D) \citep{lee2019v299} backbone network and the BEVFormer encoder \citep{li2024bevformer}. Although the performance indicators of model C are superior to those of model D, our mAS evaluation indicates that the stability of model C is relatively poor. By visually comparing the outputs of the two, we find that model C occasionally detects non-existent pedestrian crossings in individual frames, which is a manifestation of poor field stability. This observation result confirms that the stability of model C is indeed weaker than that of model D, which is consistent with the mAS evaluation result.

%----E，F-------
As depicted in Figure \ref{fig:mAP_vs_mAS_fig1}, model E represents the HRMapNet \citep{zhang2024hrmapnet} model with a training map as initial map, employing a ResNet-50 backbone \citep{he2016resnet} and BevFormer encoder \citep{li2024bevformer}, trained for 24 epochs. Model F represents the GeMap model \citep{zhang2024gemap}, employing a Swin-T backbone \citep{liu2021swintransformer} and BevFormer encoder \citep{li2024bevformer}, trained for 110 epochs. Model E achieves a higher mAP value than model F, yet according to our mAS metric evaluation, model E exhibits inferior stability compared to model F. A visual comparison of their inference results reveals that the crosswalks predicted by model E show more pronounced geometric jitter, while other instances remain similar between the two models. This observation confirms that model E's stability is indeed poorer than Model F's, consistent with the assessment provided by the mAS metric.

% ----G，H----
As can be seen from Figure \ref{fig:mAP_vs_mAS_fig2}, both model G and model H represent GeMap \citep{zhang2024gemap} models. Model G was trained for 110 epochs using the Swin-T backbone network \citep{liu2021swintransformer} and the BEVFormer encoder \citep{li2024bevformer}, while model H was trained using the V2-99 (DD3D) backbone network \citep{lee2019v299} and the BEVFormer encoder \citep{li2024bevformer}. It was also trained for 110 epochs. Although the performance indicators of model G are superior to those of model H, our mAS evaluation indicates that the stability of model G is relatively poor. Through visual comparison of the outputs of the two, we find that model G has significant spatial offset and morphological fluctuation in the prediction of road boundary lines in consecutive frames, which is a manifestation of poor field stability. This observation result confirms that the stability of Model G is indeed weaker than that of model H, which is consistent with the mAS evaluation result.

%-------I J---------
As illustrated in Figure \ref{fig:mAP_vs_mAS_fig2}, both model I and model J represent GeMap models \citep{zhang2024gemap} . Model I was trained for 110 epochs using the Swin-T backbone network \citep{liu2021swintransformer} and the BEVFormer encoder \citep{li2024bevformer}. Model J was trained for 110 epochs using the V2-99 (DD3D) backbone network \citep{lee2019v299} and the BEVFormer encoder \citep{li2024bevformer}. Model I achieves a higher mAP value than model J, yet according to our mAS metric evaluation, model I exhibits inferior stability compared to model J. Model J demonstrates superior delineation in the demarcated regions, with map instances exhibiting more precise spatial localization, while maintaining comparable performance to other models in remaining areas. This observed enhancement in output quality confirms model J's higher stability, which aligns consistently with the mAS evaluation results.

% ----K，L---
As shown in Figure \ref{fig:mAP_vs_mAS_fig2}, model K represents the MapTR \citep{liao2022maptr} model integrated with temporal features, employing a ResNet-50 backbone \citep{he2016resnet} and GKT encoder \citep{chen2022GKT}, trained for 24 epochs. Model L represent the MapTR \cite{liao2022maptr} model without temporal features, utilizing the same ResNet-50 backbone \citep{he2016resnet} and GKT encoder \citep{chen2022GKT}, also trained for 24 epochs. Model K achieves a higher mAP value than model L, yet according to our mAS metric evaluation, model K exhibits inferior stability compared to model L. Model L produces clearer map results in the demarcated areas with more accurate spatial positioning of map instances, while maintaining similar performance to other models in remaining regions. This demonstrates Model L's superior stability, which is consistent with the mAS evaluation outcomes.

% model C ：gemap_swint_bevformer_110e
% model D ：gemap_vov99_dd3d_bevformer_110e
% model E ：gemap_swint_bevformer_110e
% model F ：hrmapnet_maptrv2_nusc_r50_24ep_trainmap
% model G ：gemap_swint_bevformer_110e
% model H ：gemap_vov99_dd3d_bevformer_110e
% model I ：gemap_swint_bevformer_110e
% model J ：gemap_vov99_dd3d_bevformer_110e
% model K ：maptr_r50_gkt_24e
% model L ：maptr_r50_gkt_24e_t4

%\subsection{Feature and Query Visualization in Typical Scenarios}

\clearpage

% \section{In-depth Discussion and Analysis}
% \label{sec:appendix_8_discussion}

\section{Limitations and Future Work}
\label{sec:appendix_9_limit_future}

This study presents the first dedicated benchmark for temporal stability evaluation in online HD mapping, yet several limitations indicate directions for future research. The current benchmark is constrained by the scope of existing datasets, particularly in representing complex real-world scenarios. Our evaluation primarily relies on standard driving sequences from the nuScenes dataset \citep{caesar2020nuscenes}, which lacks systematic coverage of challenging conditions such as extreme weather, adverse illumination, and intentional adversarial scenarios. Consequently, the current assessment may not fully reflect model stability under critical edge cases that are essential for safe autonomous driving.

Another limitation stems from the rapid evolution of this research field. While our benchmark encompasses 42 model variants representing major architectural paradigms, new methodologies continue to emerge at a rapid pace. The current static snapshot of model comparisons requires continuous updates to maintain relevance and comprehensiveness.

To address these limitations, we outline two primary directions for future work. First, we will establish a continuously maintained benchmark platform that systematically incorporates new research developments. This living benchmark will implement standardized evaluation protocols for emerging methodologies, ensuring fair comparisons and tracking progress over time. The platform will feature regular updates to model implementations, evaluation metrics, and dataset expansions, fostering community-wide collaboration and providing a reliable foundation for assessing advancements in temporal stability.

Second, we will expand the benchmark to include diverse challenging scenarios that better reflect real-world complexity. This expansion will incorporate data from multiple geographic regions with varying road infrastructures and traffic patterns. Specifically, we will integrate specialized datasets containing extreme weather conditions (heavy rain, snow, fog), low-light and night-time driving scenarios, and challenging urban environments with complex intersections and dense traffic. Furthermore, we will develop evaluation protocols for synthetic adversarial scenarios designed to stress-test model stability, such as sensor degradation simulations and challenging weather transitions. These enhancements will provide a more comprehensive assessment of model robustness under critical conditions.

We believe these efforts will significantly advance the development of more reliable and robust online HD mapping systems.

\end{document}